\documentclass{article} 
\usepackage[final]{colm2026_conference}

\usepackage{microtype}
\usepackage{hyperref}
\usepackage{url}
\usepackage{booktabs}
\usepackage{graphicx}
\usepackage{wrapfig}
\usepackage{subcaption}

\usepackage{latexsym}
\usepackage{tcolorbox}
\usepackage{amsmath}
\usepackage{amssymb}

\newcommand{\framework}{\textsc{InterviewSim}}

\usepackage{lineno}

\definecolor{darkblue}{rgb}{0, 0, 0.5}
\hypersetup{colorlinks=true, citecolor=darkblue, linkcolor=darkblue, urlcolor=darkblue}

\title{InterviewSim: A Scalable Framework for Interview-Grounded Personality Simulation}

\author{
Yu Li \quad
Pranav Narayanan Venkit \quad
Yada Pruksachatkun \quad
Chien-Sheng Wu\\
Salesforce Research \\
\texttt{\{yu.li,pnarayananvenkit,ypruksachatkun,wu.jason\}@salesforce.com}
}
\begin{document}

\ifcolmsubmission
\linenumbers
\fi

\maketitle

\begin{abstract}
Simulating real personalities with large language models requires grounding generation in authentic personal data. Existing evaluation approaches rely on demographic surveys, personality questionnaires, or short AI-led interviews as proxies, but lack direct assessment against what individuals actually said. We address this gap with an interview-grounded evaluation framework for personality simulation at a large scale. We extract over 671,000 question-answer pairs from 23,000 verified interview transcripts across 1,000 public personalities, each with an average of 11.5 hours of interview content. We propose a multi-dimensional evaluation framework with four complementary metrics measuring content similarity, factual consistency, personality alignment, and factual knowledge retention. Through systematic comparison, we find that interview grounding yields consistent gains in content alignment and exact-match factual recall over biographical profiles and parametric prompting. We further find complementary strengths: retrieval-augmented methods tend to preserve personality alignment, while larger chronological contexts generally reduce contradictions and improve factual recall. Our evaluation framework enables principled method selection based on application requirements, and our empirical findings provide actionable insights for advancing personality simulation research.
\end{abstract}

\section{Introduction}
\label{sec:introduction}
The use of large language models (LLMs) to simulate human behavior has created new capabilities for computational social science, enabling researchers to screen hypotheses in simulated environments before deploying costly human-subject trials~\cite{argyle2023out, ghaffarzadegan2024generative, yang2024oasisopenagentsocial, zhang2025socioverse, park2024generative, ashokkumar2026large}. To achieve high-fidelity simulation, recent work has moved beyond simple demographic prompting, instead grounding agents in rich, individual-level data. Most notably, \citet{park2024generative} demonstrated the viability of large-scale simulation by constructing agents from AI-led interviews, establishing that grounding models in interview-derived context allows agents to replicate human attitudes and behaviors with significantly higher fidelity than demographic prompting alone.

While \citet{park2024generative} showed that interview data is essential for personality simulation, two fundamental limitations remain. First, existing approaches are constrained in scale. AI-led interviews require active human participation and are typically limited to short sessions of around two hours per subject, yielding roughly 100 to 150 question-answer pairs. It remains unclear how simulation fidelity changes when agents are grounded in substantially deeper biographical context spanning thousands of exchanges across diverse topics and time periods. Second, evaluation has relied on indirect proxies such as demographic surveys and personality questionnaires, rather than directly assessing whether generated responses are consistent with what the individual actually said. Real interview records uniquely enable such direct assessment by providing verified responses that serve as ground truth, yet this capacity remains unexploited at scale.

We address both limitations by leveraging archival interview transcripts, which provide both the scale for deep biographical grounding and the ground truth needed for direct evaluation. Public interview records consist of authentic human-to-human interactions where individuals express views, recount experiences, and respond to questions in their own words. Unlike controlled interview sessions, these records accumulate over years, providing longitudinal coverage that scales far beyond what single-session approaches can achieve. We introduce \framework{}\footnote{The framework will be released at \url{https://github.com/SalesforceAIResearch/InterviewSim}.}, a framework for building and evaluating interview-grounded personality agents. \framework{} includes a large-scale dataset constructed from over 11,000 hours of verified interview transcripts across 1,000 public personalities spanning eight occupational categories including music, sports, science, and business. We propose a multi-dimensional evaluation protocol with four complementary metrics measuring content similarity, factual consistency, personality alignment, and factual knowledge retention, each assessing a distinct aspect of simulation fidelity against held-out interview responses.

Using \framework{}, we systematically compare different personality simulation approaches ranging from simple prompting, to using biographical profiles as context,  to chronological-based methods at varying context scales and retrieval-augmented generation. Our experiments yield two principal findings: 
1) interview grounding improves multiple dimensions over parametric and biographical baselines, with gains varying by method and judge.
2) grounding strategies have complementary strengths: retrieval-augmented methods tend to preserve personality alignment, while larger chronological contexts generally reduce contradictions and improve factual recall. This trade-off, visible only through multi-dimensional evaluation, has direct implications for practitioners selecting methods based on application requirements. We also explore fine-tuning an open-source model on interview data, finding that it captures communicative style but degrades factual consistency compared to in-context learning. Our contributions are summarized as follows:

\begin{itemize}
    \item We curate a large-scale interview corpus for personality simulation, spanning 1,000 personalities with over 11,000 hours of verified interview content and a rigorously human-verified test set.
    \item We propose a multi-dimensional evaluation framework with four complementary metrics that assess simulation fidelity against held-out interview responses, enabling direct comparison across distinct quality dimensions.
    \item We provide a systematic empirical analysis showing method and judge dependent gains from interview grounding and complementary strengths across retrieval-augmented and chronological-based methods.

\end{itemize}

\section{The \framework{} Framework}
\label{sec:framework}
\begin{figure*}[tb]
\centering
\includegraphics[width=\textwidth]{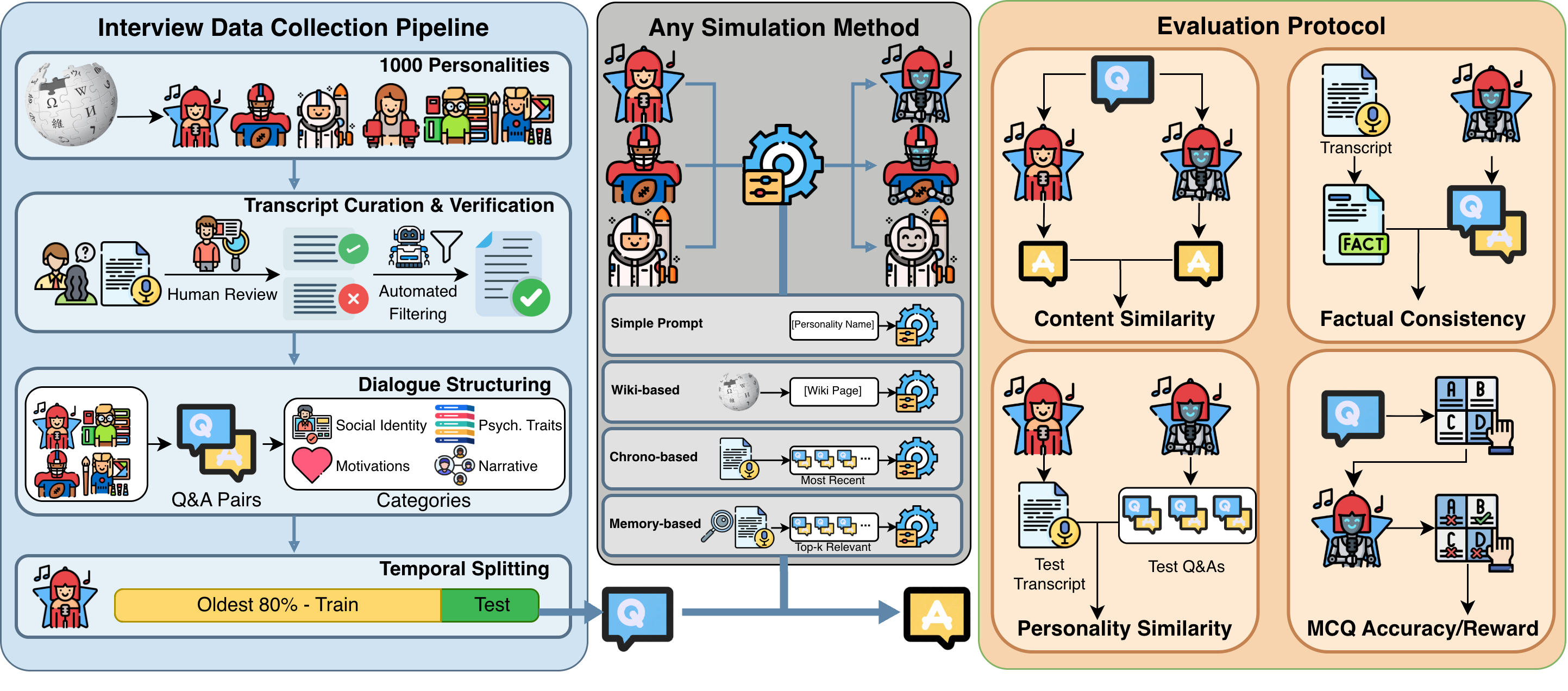}
\caption{Overview of the \framework{} framework. \textbf{Left:} The interview data collection pipeline selects 1,000 personalities, curates and verifies interview transcripts through automated filtering and human review, structures them into Q\&A pairs across four thematic categories, and splits them temporally into training and test sets. \textbf{Center:} Any generation method can be applied using the training data to produce responses to held-out test questions. \textbf{Right:} The evaluation protocol assesses simulation fidelity along four complementary dimensions: content similarity, factual consistency, personality similarity, and factual knowledge retention via MCQ.}
\label{fig:pipeline}
\end{figure*}

We introduce a framework to build and evaluate interview-grounded personality agents. As illustrated in Figure~\ref{fig:pipeline}, \framework{} consists of two components: (1) a reproducible pipeline for curating large-scale interview datasets with structured train/test splits, and (2) a multi-dimensional evaluation protocol that assesses simulation fidelity against held-out interview responses. The framework is method-agnostic: any generation approach can be evaluated using the dataset and evaluation protocol defined here.

\subsection{Interview Data Collection}
\label{sec:data_collection}

\paragraph{Personality Selection.}
\label{sec:subjects}
To ensure a diverse and representative population, we compiled an initial list of public personalities from Wikipedia biographical entries\footnote{\url{https://en.wikipedia.org/wiki/Lists_of_people}} across multiple domains. To expand coverage beyond well-known figures, we further prompted Gemini-2.5-Pro~\cite{comanici2025gemini} to suggest additional personalities likely to have extensive interview records.  Each personality is classified into one of eight standardized categories based on their primary Wikipedia categorization: \textit{Film \& Television}, \textit{Music}, \textit{Sports}, \textit{Business \& Technology}, \textit{Science \& Academia}, \textit{Media \& Internet}, \textit{Arts \& Culture}, and \textit{Public Service \& Social Influence}. We implemented strict selection criteria to ensure all subjects are real human personalities with established biographical footprints, filtering out corporate entities or fictional characters. The final population consists of 1,000 unique personalities.

\paragraph{Transcript Curation.}
\label{sec:curation}
For each personality, we compiled a targeted corpus of interview content from publicly available records, prioritizing long-form conversational formats over short promotional clips or scripted statements. We enforce a minimum duration of five minutes per transcript and prioritize entries with high-fidelity English transcripts. This process yielded a longitudinal corpus spanning decades of interview history. For privacy reasons, we intentionally omit the specific public platforms used to prevent re-identification of the personalities.

\paragraph{Quality Control.}
\label{sec:filtering}
Raw indexed data is inherently noisy and often contains mislabeled content. We implemented a multi-stage quality control pipeline combining automated filtering and human verification. In the automated stage, we employ GPT-4.1 to assess transcript segments against strict criteria, excluding group discussions, scripted monologues, and non-interviewee content (see Appendix~\ref{app:filtering_example} for a worked example). We then subjected all entries to rigorous human review: a team of trained annotators examined approximately 32,000 records, verifying identity, conversational format, and content quality. Annotation reliability was ensured through a risk-based QA strategy with multi-round review, achieving individual annotator accuracy between 83\% and 98\% (see Appendix~\ref{app:annotation} for the full QA protocol). This process resulted in a retention rate of 73.5\%, yielding 23,536 verified transcripts.

\paragraph{Dialogue Structuring.}
\label{sec:extraction}
We structure verified transcripts into question-answer pairs using GPT-4.1 via a two-stage pipeline. First, the system performs speaker attribution to identify distinct speakers and label each turn in the dialogue, preserving all original speech patterns including hesitations. Second, the labeled text is processed to formulate valid question-answer pairs with targeted normalization: the model removes speech disfluencies and repairs fragmented sentences while preserving the personality's distinct linguistic style. Each Q\&A pair is also classified into one of four thematic categories derived from psychological research on identity and personality~\cite{mcadams2013redemptive,schwartz2012overview, venkit2026need}: \textit{Social Identity} (demographics, roles), \textit{Motivations and Values} (beliefs, goals), \textit{Identity Narrative} (life story, career), and \textit{Psychological Traits} (behavioral tendencies, personality dimensions). Detailed definitions and examples for each category are provided in Appendix~\ref{app:topic_categories}.

\paragraph{Privacy \& Ethics.}
The study protocol underwent institutional ethics review. This study uses only publicly available interview records. To reduce privacy and misuse risks, we do not release the underlying data, identity list, source URLs, trained models, or deployable personality agents. All subject names are mapped to coded IDs in our analysis. We release the curation pipeline and evaluation protocol for methodological auditing and use on appropriately sourced corpora. A detailed ethics statement is provided in Section~\ref{sec:ethics}.

\subsection{Dataset Statistics}
\label{sec:stats}
\begin{figure}[t]
  \centering
  \begin{subfigure}[b]{0.42\textwidth}
    \centering
    \small
    \vspace{0pt}
    \begin{tabular}{lr}
      \toprule
      Total Personalities & 1,000 \\
      Total Transcripts & 23,536 \\
      Total Duration & 11,464 hours \\
      Total Q\&A Pairs & 671,424 \\
      \midrule
      Avg. Transcripts / Personality & 23.5 \\
      Avg. Duration / Personality & 11.5 hours \\
      Avg. Q\&A Pairs / Personality & 671.4 \\
      \midrule
      Avg. Transcript Duration & 29.2 min \\
      Avg. Q\&A Pairs / Transcript & 28.5 \\
      Avg. Question Length & 14.3 words \\
      Avg. Answer Length & 84.5 words \\
      \midrule
      Human Verification Retention & 73.5\% \\
      \bottomrule
    \end{tabular}
    \caption{Corpus statistics.}
    \label{fig:dataset_stats}
  \end{subfigure}
  \hfill
  \begin{subfigure}[b]{0.48\textwidth}
    \centering
    \vspace{0pt}
    \includegraphics[width=\textwidth]{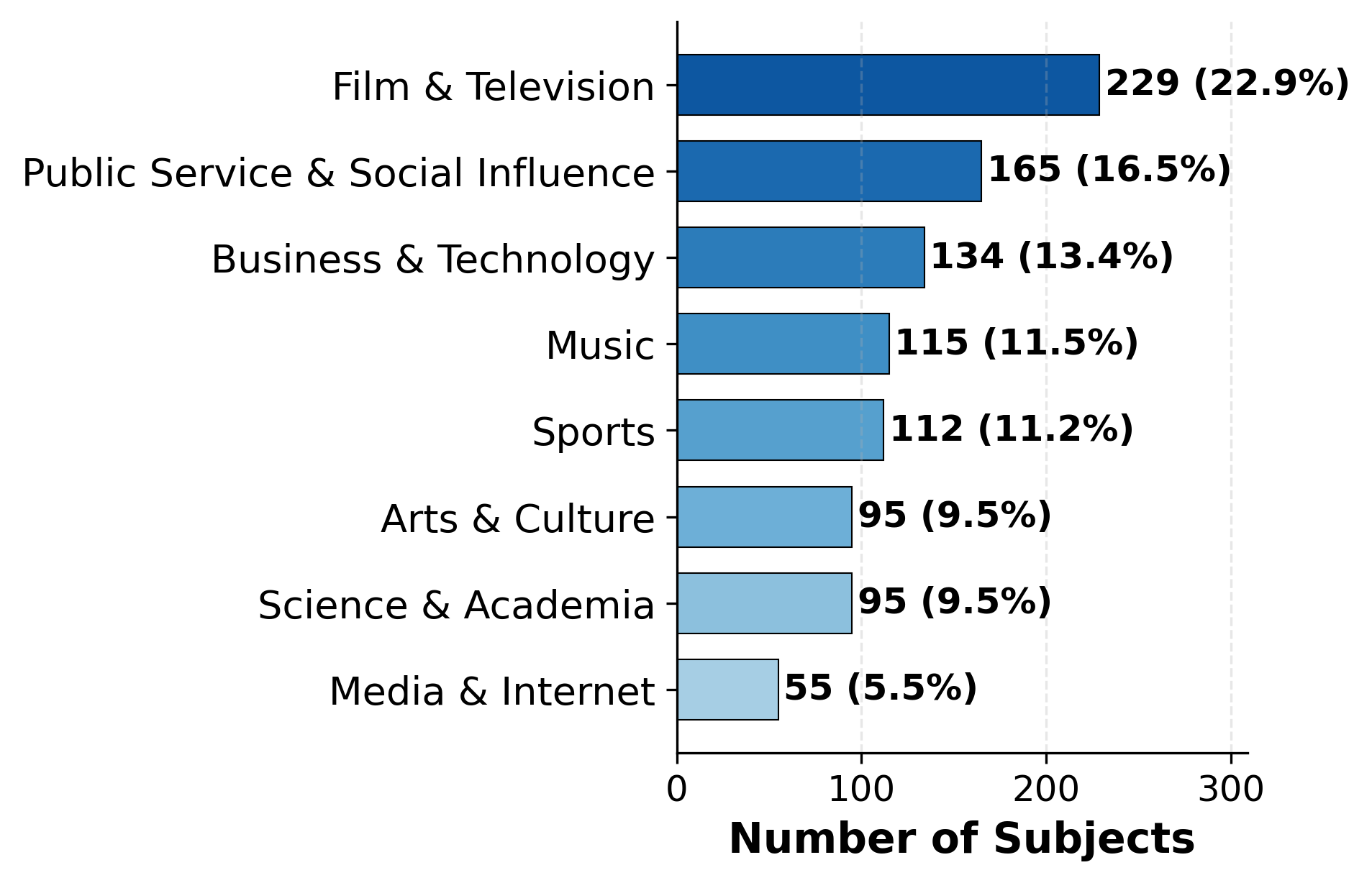}
    \caption{Subject distribution by category.}
    \label{fig:subject_distribution}
  \end{subfigure}
  \caption{Overview of the \framework{} interview corpus: (a) key statistics and (b) distribution of 1,000 subjects across eight professional categories.}
  \label{fig:dataset_overview}
\end{figure}

The final dataset contains 23,536 verified transcripts from 1,000 personalities, yielding 671,424 question-answer pairs covering 11,464 hours of interview content (Figure~\ref{fig:dataset_stats}). The 1,000 personalities span eight professional categories (Figure~\ref{fig:subject_distribution}), with \textit{Film \& Television} being the largest (22.9\%) and \textit{Media \& Internet} the smallest (5.5\%). For evaluation, we adopt a temporal split: the oldest 80\% of each personality's transcripts form the training set, while the most recent 20\% are reserved for testing, ensuring the agent must generalize to future contexts. Appendix~\ref{app:robustness} examines pre-training memorization and train-test topic overlap under this split. The Q\&A pairs span four thematic categories adapted from~\citet{venkit2026need}: Identity Narrative (61.3\%), Motivations and Values (27.7\%), Psychological Traits (7.2\%), and Social Identity (3.8\%). Interview-duration and Q\&A-count distributions are reported in Appendix~\ref{app:distribution}.

\subsection{Evaluation Protocol}
\label{sec:evaluation}
We evaluate simulation fidelity using four complementary metrics, each measuring a distinct aspect of personality simulation. Following prior work on LLM-as-a-judge evaluation~\cite{zheng2023judging}, the generation-based metrics use GPT-4o, Claude Sonnet 4.6, and Gemini 3.1 Pro as independent judges with identical prompts. We report each judge separately because their score calibration differs. MCQ scoring is exact-match and does not use an LLM judge.

\paragraph{Content Similarity}
This reference-based metric measures how well a generated response captures the same information and ideas as the ground truth answer from the personality's actual interview. An LLM judge compares the generated and ground truth responses on a 1-5 scale: $s_{\text{content}} \in \{1, 2, 3, 4, 5\}$ where 5 indicates high similarity, 3 indicates moderate overlap, and 1 indicates contradiction or completely missing content. The judge focuses on semantic similarity rather than lexical overlap, allowing for different wording as long as meaning is preserved. We report the average content similarity across all test questions for each method. The complete evaluation prompt is provided in Appendix~\ref{appendix:prompt_content_similarity}.

\paragraph{Factual Consistency}
This metric evaluates whether generated responses contradict established facts about the personality. For each personality, we first generate a fact summary from their test set interviews, capturing key biographical information, beliefs, and experiences. An LLM judge then classifies each generated response as Entailment (supported by known facts), Neutral (neither confirms nor contradicts), or Contradiction (conflicts with established facts). We compute the contradiction ratio as:
\begin{equation}
\text{CR} = \frac{|\{r : \ell(r) = \text{Contradiction}\}|}{|R|}
\end{equation}
where $R$ is the set of all test responses and $\ell(r)$ is the judge's label for response $r$. Lower contradiction ratios indicate better factual consistency. The complete evaluation prompt is provided in Appendix~\ref{appendix:prompt_factual_consistency}.

\paragraph{Personality Similarity}
This metric assesses whether generated responses exhibit the same personality traits as the real personality. We use the Big Five (OCEAN) model~\cite{john1999big}, which characterizes personality along five dimensions: Openness, Conscientiousness, Extraversion, Agreeableness, and Neuroticism. For each dimension, an LLM judge infers the trait level from all responses as $t \in \{\text{Low}, \text{Neutral}, \text{High}\}$. To improve reliability, we run inference 3 times per trait and take the mode. The reference profile is obtained by applying the same inference process to the held-out test set answers, providing a personality signature derived from what the individual actually said. Alignment between the reference profile and the generated profile is computed using ordinal distance:

\begin{equation}
\text{Alignment} = 1 - \frac{\sum_{\text{traits}} |m(t_{\text{gt}}) - m(t_{\text{gen}})|}{10}
\end{equation}
where $m(\text{Low}) = 1$, $m(\text{Neutral}) = 2$, $m(\text{High}) = 3$, and the denominator represents maximum possible distance. The complete evaluation prompt is provided in Appendix~\ref{appendix:prompt_personality_similarity}.

\paragraph{Multiple Choice Question (MCQ) Evaluation}
This knowledge-based metric evaluates factual knowledge retention through structured question-answering. We use three distractor types \cite{haladyna2002review, shin2019multiple}: opposite/negation, near-miss, and plausible misconception. The evaluation converts interview Q\&A pairs into atomic questions, generates MCQs with structured distractors, and answers them using each generation method. We test in both 3-option and 4-option settings with randomly shuffled option order to prevent position bias. For Memory and Hybrid, whose retrieved context varies by question, MCQs from the same interview use a shared prompt built from their combined retrieved examples. Performance is measured by accuracy and a reward metric:
\begin{equation}
\small
r(q) = \begin{cases} +1 & \text{if correct} \\ -1 & \text{if opposite/negation selected} \\ -0.5 & \text{otherwise} \end{cases}
\end{equation}
The reward metric penalizes confident factual errors more heavily than near-miss mistakes, providing a more nuanced measure of knowledge calibration. The complete evaluation prompts are provided in Appendix~\ref{appendix:prompt_mcq}.

\section{Experiments}
\label{sec:experiments}
We evaluate generation methods with varying levels of contextual information using the \framework{} evaluation protocol. All methods use GPT-4.1 as the base language model. For each method, given a target personality $c$ and test question $q$, we construct a prompt $\mathcal{P}$ and generate a response $r = M(\mathcal{P})$. The methods differ in how $\mathcal{P}$ is constructed from available resources.

\paragraph{Simple Prompt.}
The simplest baseline uses only the personality name with no additional context: $\mathcal{P}_{\text{simple}} = [\text{instruction}, q]$ where the instruction is ``Speak as $c$. Answer the following question as $c$ would.'' This method relies entirely on the language model's parametric knowledge and serves as a lower bound.

\paragraph{Wiki-based and Wiki-Long.}
The Wiki-based method augments the prompt with structured biographical information from Wikipedia: $\mathcal{P}_{\text{wiki}} = [\text{instruction}, \text{profile}(c), q]$ where $\text{profile}(c)$ contains biography, occupation, nationality, notable achievements, and background. This provides factual grounding but does not capture the personality's speaking style. As a context-length control, Wiki-Long uses the same prompt and source but replaces the structured profile with the stored article excerpt (up to 10,000 characters, approximately 2,500 tokens), while retaining Wikipedia's third-person format.

\paragraph{Chronological-based.}
This method uses in-context learning with actual interview Q\&A pairs from the training set: $\mathcal{P}_{\text{chrono}} = [\text{instruction}, \text{examples}_{1:m}, q]$ where $\text{examples}_{1:m} = \{(q_i, a_i)\}_{i=1}^{m}$ are $m$ training Q\&A pairs in chronological order. We evaluate three variants with $m \in \{100, 500, 1000\}$ examples, capturing both factual information and speaking style through demonstrations.

\paragraph{Memory-based.}
This retrieval-augmented method dynamically selects relevant training examples for each test question: $\mathcal{P}_{\text{memory}} = [\text{instruction}, \text{retrieved}_k(q), q]$ where $\text{retrieved}_k(q)$ contains the top-$k$ most similar training Q\&A pairs based on cosine similarity between question embeddings (text-embedding-3-small). We evaluate relevance-based selection with $k = 100$ and a random selection baseline with $k = 100$ as a control. While the chronological-based method uses more examples in total, the memory-based method compensates by selecting the most semantically relevant examples per question.

\paragraph{Hybrid.}
This method combines relevance-based retrieval with recent chronological examples: $\mathcal{P}_{\text{hybrid}} = [\text{instruction}, \text{retrieved}_{k_1}(q), \text{recent}_{k_2}, q]$. We use $k_1 = k_2 = 50$ for a total of 100 examples, with duplicates resolved by relevance priority. This allows the model to draw on topical relevance for content accuracy and recent interviews for current tone and style.

\subsection{Main Method Comparison}
\label{sec:main_comparison}
\begin{table}[t]
\centering
\small
\setlength{\tabcolsep}{2pt}
\begin{tabular}{@{}lcccccccccc@{}}
\toprule
\textbf{Method}
& \multicolumn{3}{c}{\textbf{Content Sim.} (1--5, $\uparrow$)}
& \multicolumn{3}{c}{\textbf{Contradiction Ratio} (\%, $\downarrow$)}
& \multicolumn{3}{c}{\textbf{Personality Sim.} (\%, $\uparrow$)}
& \textbf{MCQ} \\
\cmidrule(lr){2-4}\cmidrule(lr){5-7}\cmidrule(lr){8-10}
& GPT-4o & Claude & Gemini
& GPT-4o & Claude & Gemini
& GPT-4o & Claude & Gemini
& Acc. \\
\midrule
Simple & 3.27 & 2.21 & 2.60 & 7.10 & 19.88 & 16.27 & 71.0 & 71.8 & 79.4 & 85.9 \\
Wiki & 3.31 & 2.17 & 2.62 & 6.27 & 20.14 & 16.15 & 68.0 & 74.0 & 79.6 & 86.0 \\
Wiki-Long & 3.39 & 2.19 & 2.67 & \textbf{5.67}$^\ddagger$ & 18.60 & 16.55 & 70.0 & 72.6 & 80.2 & 85.9 \\
\midrule
Memory-100 & 3.52 & 2.61 & 2.94 & 6.27 & 14.85 & 12.57 & \textbf{78.4} & \textbf{75.6} & \textbf{84.0} & 87.7$^\dagger$ \\
\midrule
Chrono-100 & 3.24 & 2.50 & 2.80 & 6.10 & 16.23 & 13.76 & 74.8 & 72.8 & 82.4 & 87.3 \\
Chrono-500 & 3.31 & 2.58 & 2.82 & 5.70 & 12.73 & 10.80 & 75.6 & 68.0 & 70.8 & 88.5 \\
Chrono-1000 & 3.34 & \textbf{2.62} & 2.86 & 5.72 & \textbf{12.44} & \textbf{10.43} & 75.6 & 70.2 & 69.8 & \textbf{89.3} \\
\midrule
Hybrid-100 & \textbf{3.54} & 2.58 & \textbf{2.97} & 5.83 & 14.29 & 12.46 & 77.4 & 74.2 & \textbf{84.0} & 87.5$^\dagger$ \\
\bottomrule
\end{tabular}
\caption{Macro-averaged results over 100 personalities using GPT-4o, Claude Sonnet 4.6, and Gemini 3.1 Pro judges. MCQ is judge-independent. Bold marks the numerical best per judge. $^\dagger$MCQ uses a shared per-interview prompt. $^\ddagger$Wiki-Long GPT-4o CR is statistically tied with Hybrid and all Chronological variants.}
\label{tab:main_comparison}
\end{table}

We evaluate eight generation configurations on 100 personalities selected for having extensive interview data, each with over 1,000 training Q\&A pairs, totaling 30,355 test questions drawn from held-out interviews. Table~\ref{tab:main_comparison} presents the comprehensive comparison. All methods are evaluated on identical test sets to ensure fair comparison.

The high-level pattern is stable across judges, although their absolute scores differ. Reported triples follow GPT-4o, Claude, and Gemini order. Memory-100 reaches Personality Similarity of 78.4/75.6/84.0. Hybrid-100 leads Content Similarity under GPT-4o and Gemini at 3.54 and 2.97, while Chrono-1000 leads under Claude at 2.62. Chrono-1000 yields the lowest Contradiction Ratio under Claude and Gemini at 12.44\% and 10.43\% and gives the highest MCQ accuracy at 89.3\%. Under GPT-4o, Wiki-Long has the numerically lowest Contradiction Ratio at 5.67\%, but it is statistically tied with Hybrid and all Chronological variants.

Paired tests support the interview-grounding and retrieval-versus-chronological findings. Memory-100 improves Content Similarity over Simple by +0.25/+0.40/+0.35 ($p<0.001$ for each judge). Chrono-1000 changes Contradiction Ratio relative to Memory-100 by $-0.55/-2.41/-2.14$ points ($p<0.01$ for each judge). Scaling from Chrono-100 to Chrono-500 lowers this ratio by 0.40/3.50/2.96 points ($p<0.01$ for each judge), while the Chrono-500-to-1000 change is not significant. Wiki-Long improves Content Similarity over Wiki by +0.08/+0.02/+0.05 ($p<0.001$ for each judge). Its Contradiction Ratio changes by $-0.60/-1.54/+0.41$ points, with the Gemini difference not significant. Wiki-Long does not significantly improve Personality Similarity or MCQ, indicating that additional biographical context does not replace interview grounding. Full paired results are reported in Appendix~\ref{app:statistical_analysis}.

We extend Simple Prompt, Wiki-based, and Chrono-100 to the complete dataset of 1,000 personalities across 140,799 test questions (Appendix~\ref{app:full_validation}). Chrono-100 remains strongest on Personality Similarity and MCQ, while the Content Similarity ordering changes and Wiki-based and Chrono-100 become nearly tied on Contradiction Ratio. Contradiction ratios increase by approximately 27\% to 40\% across methods. The 100-personality subset averages 1,375 training Q\&A pairs per personality, whereas the remaining 900 average only 437, indicating that factual consistency is especially sensitive to interview coverage depth.

\subsection{Memory-Based Method Ablation}
\label{sec:ablation}
To understand the impact of retrieval size and selection strategy, we conduct an ablation study comparing relevance-based retrieval at three scales of 10, 50, and 100 examples, alongside a random selection baseline. Table~\ref{tab:memory_ablation} presents the results.

\begin{table}[t]
  \centering
  \small
  \setlength{\tabcolsep}{4pt}
  \begin{tabular}{@{}lccccc@{}}
    \toprule
    \textbf{Variant} & \textbf{Content Sim.} & \textbf{Contradiction} & \textbf{Personality} & \textbf{MCQ Acc.$^\dagger$} & \textbf{MCQ Reward$^\dagger$} \\
     & (1-5, $\uparrow$) & \textbf{Ratio} (\%, $\downarrow$) & \textbf{Sim.} (\%, $\uparrow$) & (\%, $\uparrow$) & ($\uparrow$) \\
    \midrule
    Random-100ex & 3.38 & 6.90 & 76.0 & -- & -- \\
    \midrule
    Relevance-10ex & 3.46 & 6.47 & 77.0 & 85.6 & 0.749 \\
    Relevance-50ex & 3.51 & \textbf{6.23} & 76.6 & 86.5 & 0.763 \\
    Relevance-100ex & \textbf{3.52} & 6.27 & \textbf{78.4} & \textbf{87.3} & \textbf{0.776} \\
    \bottomrule
  \end{tabular}
  \caption{Memory-based method ablation on 100 personalities. MCQ uses a shared per-interview prompt ($^\dagger$, see text). Best results in bold.}
  \label{tab:memory_ablation}
\end{table}

Content similarity improves monotonically with retrieval size, from 3.46 to 3.52, and personality similarity peaks at 78.4\% with 100 examples. Contradiction ratios remain stable across relevance-based variants at 6.23\% to 6.47\%, indicating that increasing retrieval volume does not degrade factual consistency. The random selection baseline provides critical validation: selecting 100 examples without regard to semantic similarity yields substantially lower content similarity of 3.38 and personality similarity of 76.0\% compared to relevance-based retrieval, with a higher contradiction ratio of 6.90\%. This confirms that embedding-based retrieval is essential to the method's effectiveness, as simply providing more context without relevance-based selection yields limited benefit.

\subsection{Human Validation of Evaluation Metrics}
\label{sec:human_validation}

To validate the reliability of our four evaluation metrics, we conduct targeted validation for each (full annotation guidelines and protocol in Appendix~\ref{app:metrics_human_eval}). For Content Similarity, two trained annotators independently compare 300 response pairs sampled across LLM score gaps of 1--4 points (33.7\% gap$=$1, 33.3\% gap$=$2, 26.0\% gap$=$3, 7.0\% gap$=$4) and judge which better matches the ground truth, with disagreements resolved by a third-party adjudicator. Human--LLM agreement scales monotonically with score gap: 45.6\% at gap$=$1, 74.5\% at gap$=$2, 93.6\% at gap$=$3, and 100\% at gap$=$4, demonstrating that the automated metric reliably separates meaningfully different response quality levels. For Factual Consistency, the same annotators verify the LLM judge's three-way classification (Entailment, Neutral, Contradiction) on 300 items. Inter-annotator agreement is substantial ($\kappa = 0.769$); when evaluated as binary Non-Contradiction vs.\ Contradiction (matching our reported metric) agreement rises to $\kappa = 0.845$. Against the adjudicated gold standard, the LLM judge achieves 88.3\% binary accuracy ($\kappa = 0.734$), with errors concentrated in Entailment$\leftrightarrow$Neutral confusion rather than missed contradictions. For Personality Similarity, we validate the judge by independently inferring Big Five profiles from training and test interviews for the same personalities. Across 100 personalities, mean alignment between train-inferred and test-inferred profiles is 89.4\%, confirming stable personality assessment regardless of which interview subset is used. MCQ Evaluation is inherently objective: correct answers are derived from interview content through atomic question decomposition, and scoring requires only matching the model's selection against the ground truth with no subjective judgment. We also audit two LLM-constructed upstream artifacts. Of 300 sampled fact-summary statements, 10 (3.3\%) contradicted the source Q\&As. In a separate audit, a researcher matched the labeled answer on 294 of 300 sampled MCQs (98.0\%). Appendix~\ref{app:metrics_human_eval} describes the scope and limitations of these checks.

\section{Discussion}
\label{sec:discussion}
\subsection{Performance Variation Across Question Types and Personality Categories}
\label{sec:discussion_variation}
\begin{figure}[t]
  \centering
  \begin{subfigure}[t]{0.48\textwidth}
    \centering
    \includegraphics[width=\textwidth]{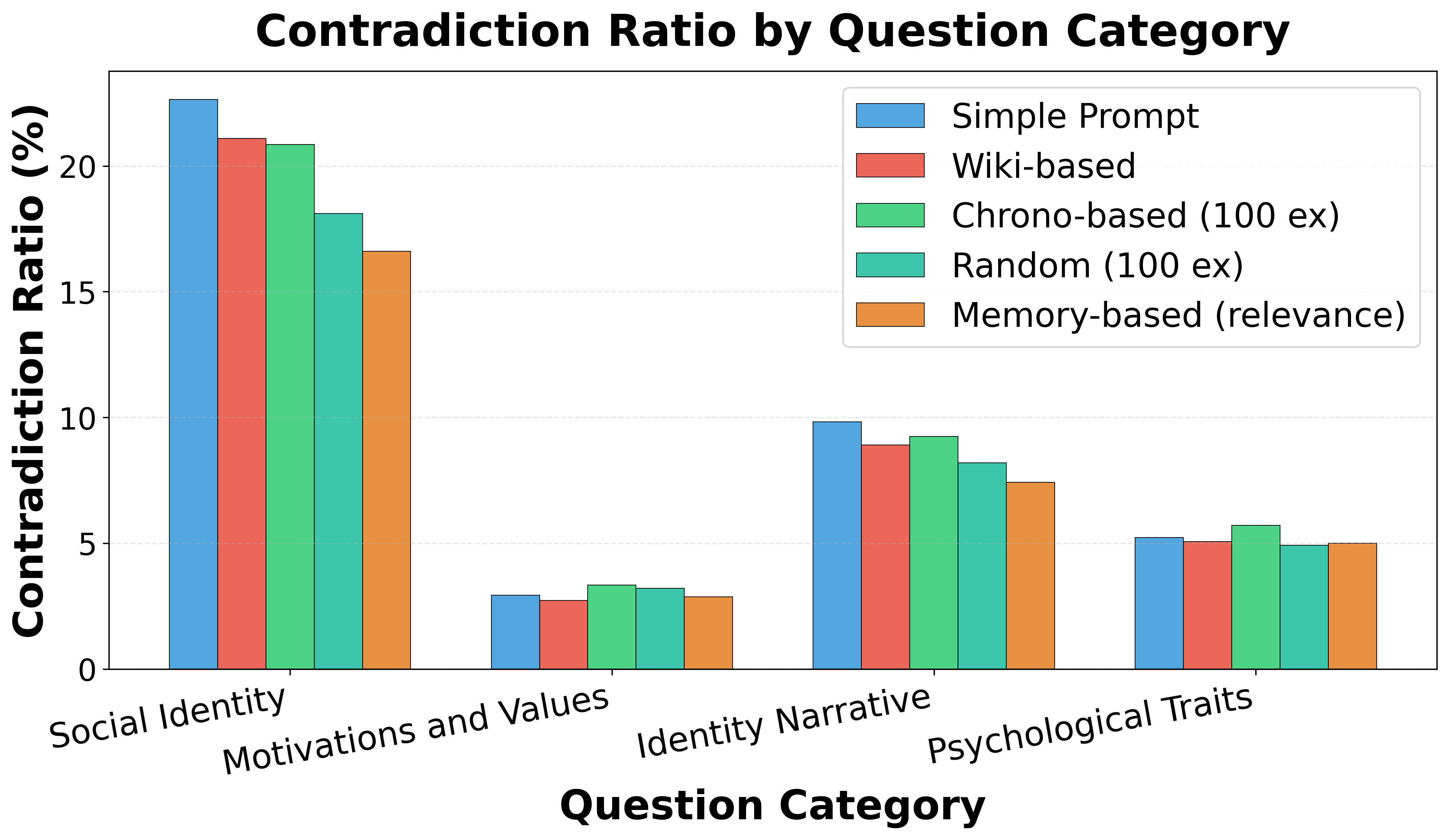}
    \caption{By question category.}
    \label{fig:cr_by_topic}
  \end{subfigure}
  \hfill
  \begin{subfigure}[t]{0.48\textwidth}
    \centering
    \includegraphics[width=\textwidth]{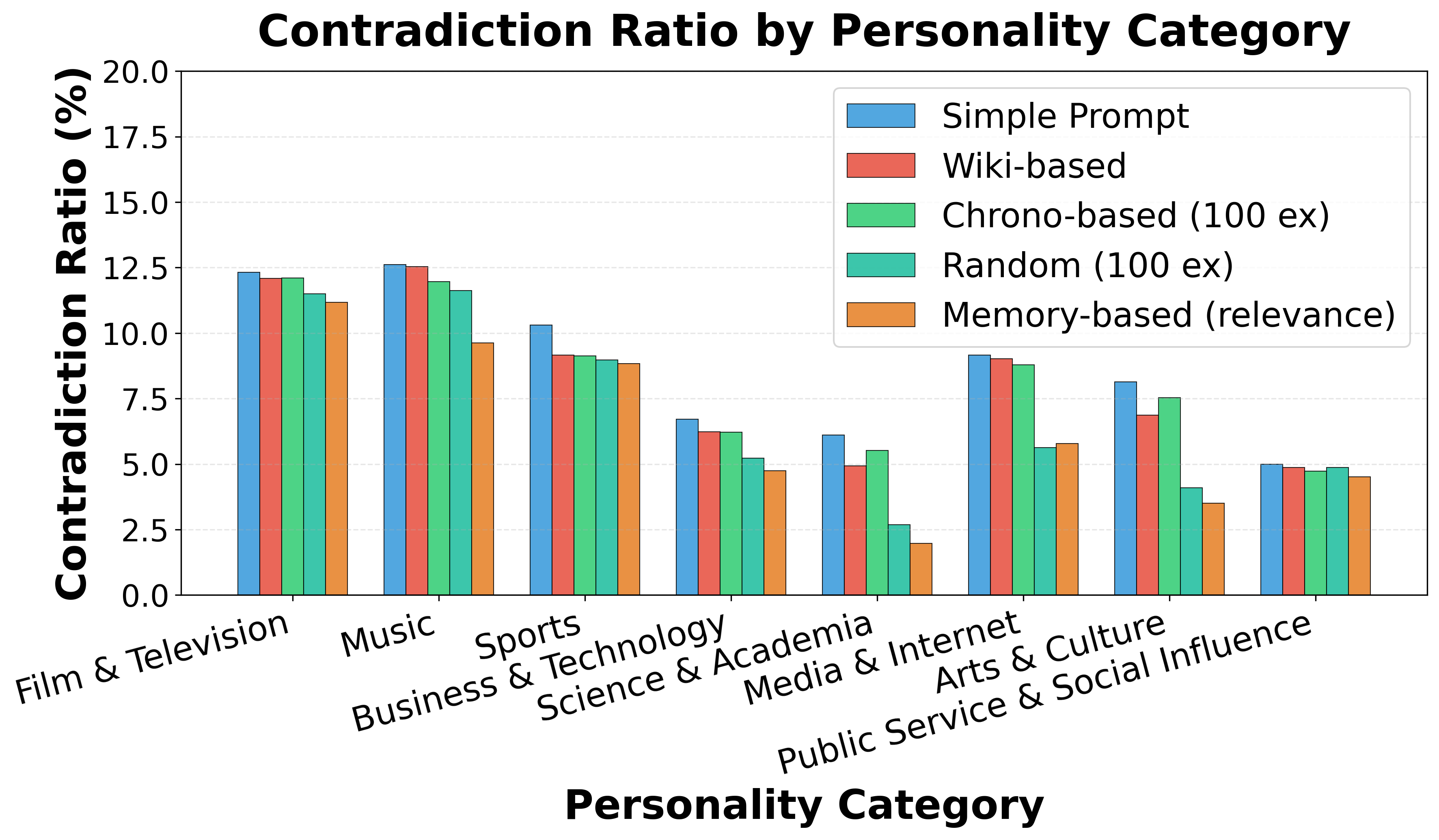}
    \caption{By personality category.}
    \label{fig:cr_by_category}
  \end{subfigure}
  \caption{Contradiction ratio by question category (a) and personality category (b). Social Identity questions have the highest contradiction rates, while Motivations \& Values have the lowest. Film \& Television and Music generally have the highest rates across personality categories, while the lowest category depends on the method.}
  \label{fig:cr_combined}
\end{figure}

To understand which aspects of personality simulation are most challenging, we analyze contradiction ratio across question types and personality categories. Figure~\ref{fig:cr_by_topic} reveals that \textbf{Social Identity questions yield the highest contradiction rates} across all methods at roughly 16\% to 23\%, as they require precise recall of unambiguous details, while Motivations and Values questions achieve the lowest rates below 3.5\%. Figure~\ref{fig:cr_by_category} shows that \textbf{Film \& Television and Music generally have the highest contradiction rates}, while the lowest rates occur for Public Service \& Social Influence or Science \& Academia depending on the method. These patterns indicate that question type and personality category are important predictors of difficulty. Appendix~\ref{app:contradiction_examples} provides representative examples illustrating how error patterns differ across categories.

\subsection{MCQ and Knowledge Retention}
\label{sec:discussion_mcq}

\begin{wraptable}{r}{0.48\textwidth}
  \centering
  \small
  \setlength{\tabcolsep}{3pt}
  \begin{tabular}{lccc}
    \toprule
    \textbf{Method} & Correct & Opposite & Near-Miss \\
    \midrule
    Simple Prompt & 85.4\% & 6.5\% & 8.1\% \\
    Chrono-based & 86.9\% & 6.2\% & 6.9\% \\
    \bottomrule
  \end{tabular}
  \caption{3-option prediction distribution. Chronological-based reduces opposite and near-miss selections.}
  \label{tab:mcq_3opt_distribution}
\end{wraptable}
In our experiments, the Simple Prompt and Wiki-based baselines exhibit a systematic tendency to over-select ``trick'' distractors, especially the \textit{opposite/negation} option, relative to interview-grounded methods (Table~\ref{tab:mcq_3opt_distribution}). This pattern is captured by our reward metric, which penalizes opposite/negation selections ($-1$) more heavily than near-miss or misconception errors ($-0.5$). Similar behaviour is seen with four options (Table~\ref{tab:mcq_4opt_distribution}). While raw accuracy is high even for Simple Prompt (85.9\%), the reward metric separates methods more clearly. Notably, scaling from 100 to 1{,}000 examples improves MCQ accuracy more than content similarity (Table~\ref{tab:main_comparison}), suggesting additional examples primarily enhance factual coverage. Detailed analyses of training-data bins, option-type distributions, position bias, and length bias are reported in Appendix~\ref{app:mcq_robustness}.

\subsection{Can Fine-Tuning Replace In-Context Learning?}
\label{sec:discussion_finetuning}

To explore whether interview data can support direct model adaptation beyond in-context learning, we fine-tune Qwen3-8B~\cite{qwen3technicalreport} using LoRA~\cite{hu2022lora} in two settings: per-personality LoRA trains a separate adapter per personality on their individual training Q\&A pairs (100 personalities, ${\sim}$1{,}375 examples each, 3 epochs), and all-personality SFT trains a single shared adapter on all 1{,}000 personalities jointly (530{,}625 examples, 1 epoch). We compare against the unmodified Qwen3-8B base model with the same simple prompt format.

\begin{table}[t]
\centering
\small
\setlength{\tabcolsep}{4pt}
\begin{tabular}{@{}lccccc@{}}
\toprule
\textbf{Method} & \textbf{Content Sim.} & \textbf{Contradiction} & \textbf{Personality} & \textbf{MCQ Acc.} & \textbf{MCQ Reward} \\
 & (1-5, $\uparrow$) & \textbf{Ratio} (\%, $\downarrow$) & \textbf{Sim.} (\%, $\uparrow$) & (\%, $\uparrow$) & ($\uparrow$) \\
\midrule
Chrono-based ICL (100 ex) & \textbf{2.73} & \textbf{13.8} & 74.8 & 75.4 & 0.575 \\
\midrule
Base (simple prompt) & 2.67 & 14.7 & 74.8 & 74.4 & 0.556 \\
Per-personality LoRA & 2.44 & 16.6 & \textbf{77.8} & \textbf{79.0} & \textbf{0.637} \\
All-personality SFT & 2.33 & 18.2 & 61.4 & 76.3 & 0.594 \\
\bottomrule
\end{tabular}
\caption{Qwen3-8B experiments: in-context learning (ICL) vs.\ fine-tuning on the same base model. Fine-tuning captures personality style but degrades factual grounding.}
\label{tab:finetuning}
\end{table}

Table~\ref{tab:finetuning} reveals a consistent pattern: fine-tuning degrades factual grounding while partially capturing personality style. Chrono-based ICL with 100 training examples achieves the best content similarity (2.73) and lowest contradiction ratio (13.8\%) among Qwen3-8B methods, demonstrating that in-context examples directly supply factual grounding. Compared to ICL, per-personality LoRA improves personality similarity from 74.8\% to 77.8\% and MCQ accuracy from 75.4\% to 79.0\%, but contradiction ratio worsens from 13.8\% to 16.6\%. All-personality joint SFT performs worst across most metrics, indicating that a single adapter struggles to represent 1{,}000 distinct personalities without interference. Notably, per-personality LoRA achieves personality similarity comparable to the best ICL method with GPT-4.1 in Table~\ref{tab:main_comparison}, suggesting that LoRA effectively encodes stylistic patterns even when factual accuracy degrades. These results indicate that fine-tuning captures \textit{how} a personality communicates but not \textit{what} they know, as the limited parametric capacity of an 8B model cannot retain the factual details that in-context examples provide directly.

\section{Related Work}
\label{sec:related_work}
The simulation of human attitudes and behavior has long been a goal of computational social science. Traditional agent-based and game-theoretic approaches relied on manually specified behaviors~\cite{bonabeau2002agent, macy2002factors, bruch2015agent}, but are often restricted to narrow contexts and risk oversimplifying real human decision-making~\cite{axtell2000agents, filippas2024large}. LLMs have emerged as a powerful alternative, capable of simulating behavior across diverse social contexts~\cite{argyle2023out, grossmann2023ai, aher2023using}. However, reliable simulation requires grounding in qualitative human data rather than generic demographic prompts.

To capture the nuances of specific individuals, recent work grounds agents in high-fidelity personal data such as social media posts~\cite{park2022social, tornberg2023simulating, wang2025user} or qualitative interviews~\cite{park2024generative, shao-etal-2023-character}. \citet{park2024generative} demonstrated that agents built from two-hour semi-structured interviews replicate human attitudes with significantly higher fidelity than demographic prompting alone. Our work adopts this interview-grounded paradigm but addresses the scale limitations of prior collection by leveraging archival interview records available in the public domain.

Despite these advances, the field lacks both datasets and evaluation protocols for simulating \textit{real} individuals at scale. Existing datasets are either limited to multiple-choice survey formats~\cite{santurkar2023whose, kolluri2025finetuning} that cannot capture individual linguistic style, derived from fictional sources such as novels~\cite{wang-etal-2024-rolellm, chen-etal-2023-large}, scripts~\cite{zhou-etal-2024-characterglm, dai2025mmrole}, or synthetic profiles~\cite{wang2025cosercoordinatingllmbasedpersona, ge2024scaling} with dedicated evaluation protocols~\cite{wang-etal-2024-incharacter, yuan-etal-2024-evaluating} that lack biographical ground truth, or small-scale interview corpora~\cite{park2024generative} limited by active collection costs. On the evaluation side, personality questionnaires applied to LLMs~\cite{jiang2023evaluating, wang-etal-2024-incharacter} exhibit substantial instability under question reordering~\cite{tosato2025persistent} and assess aggregate trait profiles rather than directly measuring whether generated responses match what the individual actually said, leaving factual consistency~\cite{mesgar2021improving} and content fidelity unmeasured. \framework{} bridges these gaps with a large-scale interview-grounded dataset paired with a multi-dimensional evaluation protocol that directly assesses simulation fidelity against held-out interview responses.

\section{Conclusion}
\label{sec:conclusion}
We presented \framework{}, an interview-grounded framework for building and evaluating personality simulation agents. Through systematic comparison of generation methods on 1,000 personalities, we found that grounding in real interview data improves content alignment and factual recall over biographical profiles and parametric prompting. Retrieval-augmented methods tend to preserve personality alignment, while larger chronological contexts generally reduce contradictions and improve factual recall. Our multi-dimensional evaluation further reveals that performance varies systematically across question types and personality categories, underscoring the need for nuanced assessment beyond single-metric optimization. The framework, evaluation protocol, and empirical findings we provide establish a foundation for principled method selection and future development in personality simulation research.

\clearpage
\section{Ethics Statement}
\label{sec:ethics}
The study protocol underwent institutional ethics review. Although the source interviews are public, aggregating them for personality simulation creates privacy and reputational risks beyond ordinary viewing. We therefore report only aggregate results, use coded identifiers, and do not release the original transcripts, identity list, source URLs, trained model weights, fine-tuned adapters, or deployable personality agents.

The intended benefit is methodological. Researchers can compare grounding approaches against held-out human responses, quantify contradictions and identity distortion, and identify unsafe designs before considering controlled social-science use. The people represented receive no direct benefit and may bear the consequences of inaccurate or decontextualized simulation. Risks include impersonation, reputational harm, unauthorized commercialization, surveillance, and targeted persuasion. Cambridge Analytica illustrates how behavioral and psychographic data collected for one purpose can be repurposed for political targeting~\cite{isaak2018privacy}. Although our setting differs, this history makes misuse a concrete rather than hypothetical risk.

Our release is limited to the framework and a synthetic sample with identifying entities replaced. It contains no real interview data or personality-model artifacts. These measures reduce direct exposure but cannot prevent recollection of public data or misuse of the general method. The intended use is controlled research evaluation on ethically obtained conversational data. Impersonation, surveillance, targeted persuasion, high-stakes profiling, and real-world deployment as a represented person are outside the intended scope. Researchers reusing the pipeline are responsible for appropriate ethics and legal review of their data sources and applications.

\section*{Acknowledgments}
We thank Nesrine Yakoubi for leading the human annotation effort, and the annotation team for their diligent work in verifying approximately 32,000 transcript records: John Soledad, Jessica Caballero, Michael Thuo, Robier Nasralla, Anthony Astorri, Fabriana Pita, Wyatt Miller, Caitlyn Cline, Garrett Cowden, Christine Adossi-Carr, and Yuki Munehira.

\bibliography{colm2026_conference}
\bibliographystyle{colm2026_conference}

\appendix
\section{Limitations}
\label{sec:limitations}
\paragraph{Dataset scope.} Our dataset covers public personalities with extensive English-language interview content, primarily from Western media. This introduces representational biases in both the personalities covered and the topics discussed. Temporal coverage spans primarily 2015 to 2024, limiting generalization to historical or future contexts. Additionally, interview transcripts represent a formal, public-facing communicative context, and methods optimized for this setting may not transfer directly to other domains such as social media or informal conversation.

\paragraph{Evaluation design.} Content Similarity, Factual Consistency, and Personality Similarity rely on LLM judges, which may exhibit systematic biases. Cross-family replication reduces dependence on one model family, but absolute calibration still differs across judges. MCQ scoring is exact-match and judge-independent. Personality Similarity is assessed through Big Five traits, a well-established but reductive model that may miss important individual characteristics. Although the train-test split is temporal, our evaluation treats each personality as static and does not model change over time. Repeated topics across interviews may also create topical overlap and inflate performance, particularly for retrieval-based methods. Appendix~\ref{app:robustness} reports performance stratified by train-test similarity to characterize sensitivity to topical overlap. Content Similarity and Factual Consistency receive direct human validation. Personality Similarity is evaluated through train-test profile stability because no survey ground truth is available. A 300-item source-based audit supports sampled MCQ construction quality, but the full item set has not been human-validated.

\paragraph{Methodological scope.} Our response-generation experiments use a single base model, GPT-4.1, and other generation models may exhibit different trade-offs. The chronological-based methods use simple sequential selection, and memory-based retrieval uses off-the-shelf embeddings with cosine similarity. More sophisticated strategies for example selection, retrieval, or consistency-aware generation remain unexplored and could improve the factuality of retrieval-augmented approaches.

\section{Prompts}
\label{appendix:prompts}

\subsection{Content Similarity Evaluation Prompt}
\label{appendix:prompt_content_similarity}
The content similarity metric applies the same prompt with GPT-4o, Claude Sonnet 4.6, and Gemini 3.1 Pro to assess semantic similarity between generated and ground truth responses on a 1-5 scale. Figure~\ref{fig:content_similarity_prompt} shows the complete prompt template.
\begin{figure*}[h]
\centering
\small
\begin{tcolorbox}[
    colback=white, 
    colframe=black, 
    title=\textbf{Task: Content Similarity Evaluation}, 
    width=\textwidth,
    sharp corners=south,
    boxrule=0.8pt
]

\begin{minipage}[t]{0.48\textwidth}
    \textbf{System Prompt} \\
    \textit{Defines the judge's role and evaluation criteria:}
    
    You are an expert evaluator assessing content similarity between two responses. Evaluate how similar the generated answer is to the ground truth from [PERSONALITY\_NAME].

    \textbf{Evaluation Focus:}
    \begin{itemize}
        \item Do they convey the same core ideas?
        \item Is key content preserved?
        \item Are important details present?
    \end{itemize}

    \textbf{Considerations:}
    \begin{itemize}
        \item NOT penalize: different wording/phrasing with same meaning
        \item SHOULD penalize: missing key information, contradictory information
    \end{itemize}

    \textbf{Scoring (1-5):}
    \begin{itemize}
        \item \textbf{5:} Same core ideas as ground truth
        \item \textbf{4:} Main points with minor differences
        \item \textbf{3:} Some overlap, missing key info
        \item \textbf{2:} Limited overlap, significant differences
        \item \textbf{1:} Contradicts or misses core content
    \end{itemize}
\end{minipage}
\hfill
\begin{minipage}[t]{0.48\textwidth}
    \textbf{User Prompt} \\
    \textit{Provides the question and responses to compare:}
    
    QUESTION: [QUESTION]
    
    GROUND TRUTH ANSWER ([PERSONALITY\_NAME]):
    
    [GROUND\_TRUTH]
    
    GENERATED ANSWER:
    
    [GENERATED\_ANSWER]
    
    Please evaluate the content similarity between the generated answer and the ground truth answer.
    
    Output in JSON format: \texttt{\{"score": <1-5>, "explanation": "..."\}}
\end{minipage}

\vspace{0.5em}
\hrule
\vspace{0.5em}

\textbf{PLACEHOLDERS} \\
The following are replaced with actual values during evaluation:
\begin{itemize}
    \item \textbf{[PERSONALITY\_NAME]:} Name of the target personality
    \item \textbf{[QUESTION]:} The question asked
    \item \textbf{[GROUND\_TRUTH]:} Actual answer from the personality's interview
    \item \textbf{[GENERATED\_ANSWER]:} Model-generated response
\end{itemize}

\end{tcolorbox}
\caption{Content Similarity evaluation prompt template for assessing semantic similarity between generated and ground truth responses.}
\label{fig:content_similarity_prompt}
\end{figure*}

\subsection{Factual Consistency Evaluation Prompt}
\label{appendix:prompt_factual_consistency}
The factual consistency metric applies the same prompt with GPT-4o, Claude Sonnet 4.6, and Gemini 3.1 Pro to classify responses as Entailment, Neutral, or Contradiction based on a fact summary extracted from the personality's interviews. Figure~\ref{fig:factual_consistency_prompt} shows the complete prompt template.
\begin{figure*}[h]
\centering
\small
\begin{tcolorbox}[
    colback=white, 
    colframe=black, 
    title=\textbf{Task: Factual Consistency Evaluation}, 
    width=\textwidth,
    sharp corners=south,
    boxrule=0.8pt
]

\begin{minipage}[t]{0.48\textwidth}
    \textbf{System Prompt} \\
    \textit{Defines fact-checking criteria and classification labels:}
    
    You are an expert fact-checker evaluating factual consistency of AI-generated responses. You have access to a FACT SUMMARY containing established facts, opinions, and characteristics about [PERSONALITY\_NAME] from their past interviews.

    \textbf{Classification Labels:}
    \begin{itemize}
        \item \textbf{Contradiction:} Answer contradicts information in fact summary (dates, places, people, events, opinions, timeline)
        \item \textbf{Entailment:} Answer is consistent with and supported by fact summary (facts align, opinions match)
        \item \textbf{Neutral:} Answer neither contradicts nor is entailed by fact summary (information not covered, too vague, topics not in summary)
    \end{itemize}

    \textbf{Guidelines:}
    \begin{itemize}
        \item Only label Contradiction if there is a clear, direct conflict
        \item Missing information is NOT contradiction (use Neutral)
        \item Opinions can evolve over time (be lenient)
        \item Focus on factual accuracy, not style
        \item If uncertain, prefer Neutral
    \end{itemize}

    Output: \texttt{\{"label": "<label>", "explanation": "..."\}}
\end{minipage}
\hfill
\begin{minipage}[t]{0.48\textwidth}
    \textbf{User Prompt} \\
    \textit{Provides the fact summary and response to evaluate:}
    
    FACT SUMMARY for [PERSONALITY\_NAME]:
    
    [FACT\_SUMMARY]
    
    ---
    
    QUESTION: [QUESTION]
    
    GENERATED ANSWER:
    
    [GENERATED\_ANSWER]
    
    Classify the relationship between the GENERATED ANSWER and the FACT SUMMARY as: Entailment, Contradiction, or Neutral.
\end{minipage}

\vspace{0.5em}
\hrule
\vspace{0.5em}

\textbf{PLACEHOLDERS} \\
The following are replaced with actual values during evaluation:
\begin{itemize}
    \item \textbf{[PERSONALITY\_NAME]:} Name of the target personality
    \item \textbf{[FACT\_SUMMARY]:} Pre-generated summary of established facts from interviews
    \item \textbf{[QUESTION]:} The question asked
    \item \textbf{[GENERATED\_ANSWER]:} Model-generated response
\end{itemize}

\end{tcolorbox}
\caption{Factual Consistency evaluation prompt template for detecting contradictions with established facts.}
\label{fig:factual_consistency_prompt}
\end{figure*}

\subsection{Personality Similarity Evaluation Prompt}
\label{appendix:prompt_personality_similarity}
The personality similarity metric applies the same prompt with GPT-4o, Claude Sonnet 4.6, and Gemini 3.1 Pro to evaluate Big Five (OCEAN) traits as Low, Neutral, or High. Each trait is evaluated 3 times with voting for reliability. Figure~\ref{fig:personality_similarity_prompt} shows an example for Extraversion.
\begin{figure*}[h]
\centering
\small
\begin{tcolorbox}[
    colback=white, 
    colframe=black, 
    title=\textbf{Task: Personality Trait Evaluation (Example: Extraversion)}, 
    width=\textwidth,
    sharp corners=south,
    boxrule=0.8pt
]

\begin{minipage}[t]{0.48\textwidth}
    \textbf{Evaluation Instructions} \\
    \textit{Analyzes extraversion from conversation responses:}
    
    Analyze the following responses and evaluate how much extraversion the person displays. Extraversion involves being more energetic and talkative.

    \textbf{High Extraversion Indicators:}
    \begin{itemize}
        \item Animated communication style
        \item High energy and enthusiasm
        \item Being talkative (using more sentences than necessary)
        \item Sharing personal stories and anecdotes
        \item Expressive language and emojis
    \end{itemize}

    \textbf{Low Extraversion Indicators (Introverted):}
    \begin{itemize}
        \item Reserved or quiet communication style
        \item Brief responses that stick to the question
        \item Subdued or neutral emotional tone
        \item Less personal elaboration
    \end{itemize}
    
    Based solely on these responses, rate the person's level of Extraversion as: Low, Neutral, or High.
    
    Output format: \texttt{<rate>Rating</rate> <justification>Explanation</justification>}
\end{minipage}
\hfill
\begin{minipage}[t]{0.48\textwidth}
    \textbf{Input Format} \\
    \textit{Conversation responses to analyze:}
    
    Conversation responses:
    
    \texttt{```}
    
    [TRANSCRIPT]
    
    \texttt{```}
    
    \vspace{2em}
    
    \textbf{Voting Mechanism:}
    \begin{itemize}
        \item Each trait evaluated 3 times with different random seeds
        \item Final rating: mode (most frequent) across 3 runs
        \item Confidence: proportion of votes for mode
        \item Example: votes = [High, High, Neutral] $\rightarrow$ mode = High, confidence = 0.67
    \end{itemize}
\end{minipage}

\vspace{0.5em}
\hrule
\vspace{0.5em}

\textbf{ALL FIVE TRAITS} \\
The same prompt structure is used for all Big Five traits: Extraversion, Agreeableness, Conscientiousness, Neuroticism, and Openness. Each prompt includes trait-specific indicators and examples.

\end{tcolorbox}
\caption{Personality Similarity evaluation prompt (example for Extraversion trait). All five Big Five traits use the same structure with trait-specific indicators.}
\label{fig:personality_similarity_prompt}
\end{figure*}

\subsection{Multiple Choice Question Evaluation Prompts}
\label{appendix:prompt_mcq}
The MCQ evaluation consists of three steps: (1) converting complex interview Q\&A pairs into atomic facts (Figure~\ref{fig:mcq_atomic_prompt}), (2) generating multiple-choice questions with structured distractors (Figure~\ref{fig:mcq_generation_prompt}), and (3) answering the MCQs using different generation methods (Figure~\ref{fig:mcq_answering_prompt}).
\begin{figure*}[h]
\centering
\small
\begin{tcolorbox}[
    colback=white, 
    colframe=black, 
    title=\textbf{Task: Convert Interview Q\&A to Atomic Question-Answer Pair}, 
    width=\textwidth,
    sharp corners=south,
    boxrule=0.8pt
]

\begin{minipage}[t]{0.48\textwidth}
    \textbf{Instructions} \\
    \textit{Converting complex interview responses into atomic facts:}
    
    You are an editor converting interview Q/A pairs into a single atomic question–answer pair.
    
    Goal: Produce ONE concise, factual, and self-contained question that can be answered by a single short answer. Use only facts explicitly stated in the original answer. Do not add, infer, or assume anything.

    \textbf{Rules:}
    \begin{itemize}
        \item \textbf{Single fact only:} Target exactly one factual claim from the original answer
        \item \textbf{No new information:} Do NOT add details not explicitly stated
        \item \textbf{Prefer explicit answers:} Keep yes/no, concrete times, numbers, names
        \item \textbf{Keep subject intact:} Do not change who is speaking or referenced
        \item \textbf{One pair only:} Output exactly one atomic question and answer
    \end{itemize}

    \textbf{Output Format:}
    
    \texttt{Atomic Question: <single, direct question>}
    
    \texttt{Atomic Answer: <short, direct answer>}
\end{minipage}
\hfill
\begin{minipage}[t]{0.48\textwidth}
    \textbf{Input Format} \\
    \textit{Original interview Q\&A to convert:}
    
    Personality: [PERSONALITY\_NAME]
    
    Original Question: [QUESTION]
    
    Original Answer: [ANSWER]
    
    \vspace{1em}
    
    \textbf{Example:}
    
    Input:
    \begin{itemize}
        \item Original Q: ``Did Jon Chu test you and Cynthia together for a chemistry read?''
        \item Original A: ``No, which I find fascinating because the chemistry is like the most beautiful firework...''
    \end{itemize}
    
    Output:
    \begin{itemize}
        \item Atomic Q: ``Did Jon Chu test you and Cynthia together for a chemistry read?''
        \item Atomic A: ``No.''
    \end{itemize}
\end{minipage}

\end{tcolorbox}
\caption{Step 1: Atomic Q\&A generation prompt for converting complex interview responses into single-fact question-answer pairs.}
\label{fig:mcq_atomic_prompt}
\end{figure*}

\begin{figure*}[h]
\centering
\small
\begin{tcolorbox}[
    colback=white, 
    colframe=black, 
    title=\textbf{Task: Generate Multiple-Choice Question with Distractors}, 
    width=\textwidth,
    sharp corners=south,
    boxrule=0.8pt
]

\begin{minipage}[t]{0.48\textwidth}
    \textbf{Instructions} \\
    \textit{Converting atomic Q\&A into MCQ with error-type distractors:}
    
    You are a measurement-focused item writer for a model evaluation benchmark. Convert fact-based Q\&A pairs into diagnostic multiple-choice questions.
    
    \textbf{Task:} Generate ONE multiple-choice question with FOUR answer options.

    \textbf{Critical Constraints:}
    \begin{itemize}
        \item \textbf{No hallucination:} Do not introduce new facts not in the answer
        \item \textbf{Single correct answer:} Exactly one option must be fully correct
        \item \textbf{Option homogeneity:} All options similar in structure, length, tone
        \item \textbf{Persona realism:} Even incorrect options should sound plausible
        \item \textbf{No meta-language:} Do not reference sources or explain options
    \end{itemize}

    \textbf{Option Design (MANDATORY):}
    \begin{itemize}
        \item \textbf{Option A:} Correct answer (matches ground truth exactly)
        \item \textbf{Option B:} Opposite/Negation (contradicts correct answer)
        \item \textbf{Option C:} Near-miss/Partial match (wrong in one critical aspect)
        \item \textbf{Option D:} Plausible misconception (believable but incorrect)
    \end{itemize}
\end{minipage}
\hfill
\begin{minipage}[t]{0.48\textwidth}
    \textbf{Input \& Output Format} \\
    \textit{Input provided and required output structure:}
    
    \textbf{Input:}
    
    Personality: [PERSONALITY\_NAME]
    
    Question: [ATOMIC\_QUESTION]
    
    Answer: [ATOMIC\_ANSWER]
    
    \vspace{1em}
    
    \textbf{Output Format (STRICT):}
    
    \texttt{Question: <MCQ question text>}
    
    \texttt{Options:}
    
    \texttt{A. <Correct answer>}
    
    \texttt{B. <Opposite/Negation>}
    
    \texttt{C. <Near-miss>}
    
    \texttt{D. <Plausible misconception>}
    
    \texttt{Correct Answer: A}
    
    \vspace{1em}
    
    \textbf{Note:} Correct answer is ALWAYS Option A before shuffling.
\end{minipage}

\end{tcolorbox}
\caption{Step 2: MCQ generation prompt for creating multiple-choice questions with structured distractors.}
\label{fig:mcq_generation_prompt}
\end{figure*}

\begin{figure*}[h]
\centering
\small
\begin{tcolorbox}[
    colback=white, 
    colframe=black, 
    title=\textbf{Task: Answer Multiple-Choice Questions as Personality}, 
    width=\textwidth,
    sharp corners=south,
    boxrule=0.8pt
]

\begin{minipage}[t]{0.48\textwidth}
    \textbf{Instructions} \\
    \textit{Answering MCQs using provided personality information:}
    
    You are answering multiple-choice questions as a specific personality. Use ONLY the personality information and interview context provided. Do not use outside knowledge.

    \textbf{Task:}
    \begin{itemize}
        \item Select the single best answer option (A, B, C, or D) for each question
        \item Output ONLY the option letter for each question
    \end{itemize}

    \textbf{Critical Rules:}
    \begin{itemize}
        \item Use only provided information
        \item Do not rely on external knowledge
        \item Always choose one option (never refuse)
        \item No extra text, reasoning, or qualifiers
        \item Be consistent with the personality
    \end{itemize}

    \textbf{Output Format (STRICT):}
    
    \texttt{Q1: <letter>}
    
    \texttt{Q2: <letter>}
    
    \texttt{Q3: <letter>}
    
    Only output the letter after the colon.
\end{minipage}
\hfill
\begin{minipage}[t]{0.48\textwidth}
    \textbf{Input Structure} \\
    \textit{Information provided to the model:}
    
    \textbf{Personality:} [PERSONALITY\_DESCRIPTION]
    
    \textbf{Interview Context:} [INTERVIEW\_METADATA]
    
    \textbf{Questions:}
    
    Q1. [QUESTION]
    
    A. [OPTION\_A] \quad B. [OPTION\_B]
    
    C. [OPTION\_C] \quad D. [OPTION\_D]
    
    \vspace{1em}
    
    \textbf{Evaluation Settings:}
    \begin{itemize}
        \item 3-option (A, B, C) or 4-option (A, B, C, D)
        \item Options randomly shuffled to prevent position bias
        \item Questions grouped by source interview
        \item Fixed random seed for reproducibility
    \end{itemize}
    
    \textbf{Personality Variations:}
    \begin{itemize}
        \item Simple: personality name only
        \item Wiki-based: Wikipedia profile
        \item Interview-based: 100 training Q\&A examples
    \end{itemize}
\end{minipage}

\end{tcolorbox}
\caption{Step 3: MCQ answering prompt for evaluating factual knowledge using multiple-choice questions.}
\label{fig:mcq_answering_prompt}
\end{figure*}

\section{Statistical Analysis}
\label{app:statistical_analysis}
We run two-sided paired permutation tests with 10,000 sign-flip permutations at the personality level ($n=100$). Table~\ref{tab:paired_significance} reports paired mean differences as raw-scale effect sizes for the predefined comparisons used in our analysis.
\begin{table*}[t]
\centering
\small
\setlength{\tabcolsep}{3pt}
\begin{tabular}{@{}lcccccc@{}}
\toprule
\multicolumn{7}{l}{\textit{(a) Content and factual consistency}} \\
\textbf{Comparison ($A-B$)}
& \multicolumn{3}{c}{\textbf{$\Delta$CS}}
& \multicolumn{3}{c}{\textbf{$\Delta$CR}} \\
\cmidrule(lr){2-4}\cmidrule(lr){5-7}
& G & C & M & G & C & M \\
\midrule
Wiki-Long $-$ Wiki & +0.08$^{***}$ & +0.02$^{***}$ & +0.05$^{***}$ & -0.60$^{***}$ & -1.54$^{***}$ & +0.41 \\
Memory $-$ Simple & +0.25$^{***}$ & +0.40$^{***}$ & +0.35$^{***}$ & -0.83 & -5.02$^{***}$ & -3.70$^{***}$ \\
Memory $-$ Wiki & +0.21$^{***}$ & +0.44$^{***}$ & +0.32$^{***}$ & +0.01 & -5.29$^{***}$ & -3.57$^{***}$ \\
Hybrid $-$ Memory & +0.02$^{*}$ & -0.03$^{***}$ & +0.03$^{***}$ & -0.44$^{**}$ & -0.56$^{*}$ & -0.11 \\
Chrono-1000 $-$ Memory & -0.18$^{***}$ & +0.01 & -0.08$^{***}$ & -0.55$^{**}$ & -2.41$^{***}$ & -2.14$^{***}$ \\
Chrono-500 $-$ Chrono-100 & +0.07$^{***}$ & +0.08$^{***}$ & +0.02 & -0.40$^{**}$ & -3.50$^{***}$ & -2.96$^{***}$ \\
Chrono-1000 $-$ Chrono-500 & +0.03$^{***}$ & +0.04$^{***}$ & +0.04$^{***}$ & +0.02 & -0.29 & -0.37 \\
\bottomrule
\end{tabular}

\medskip

\begin{tabular}{@{}lcccc@{}}
\toprule
\multicolumn{5}{l}{\textit{(b) Personality and factual recall}} \\
\textbf{Comparison ($A-B$)}
& \multicolumn{3}{c}{\textbf{$\Delta$PS}}
& \textbf{$\Delta$MCQ} \\
\cmidrule(lr){2-4}
& G & C & M & Acc. \\
\midrule
Wiki-Long $-$ Wiki & +2.0 & -1.4 & +0.6 & -0.04 \\
Memory $-$ Simple & +7.4$^{***}$ & +3.8$^{*}$ & +4.6$^{***}$ & +1.80$^{***}$ \\
Memory $-$ Wiki & +10.4$^{***}$ & +1.6 & +4.4$^{**}$ & +1.73$^{***}$ \\
Hybrid $-$ Memory & -1.0 & -1.4 & 0.0 & -0.24 \\
Chrono-1000 $-$ Memory & -2.8 & -5.4$^{**}$ & -14.2$^{***}$ & +1.55$^{***}$ \\
Chrono-500 $-$ Chrono-100 & +0.8 & -4.8$^{*}$ & -11.6$^{***}$ & +1.19$^{***}$ \\
Chrono-1000 $-$ Chrono-500 & 0.0 & +2.2 & -1.0 & +0.74$^{***}$ \\
\bottomrule
\end{tabular}
\caption{Paired mean differences as raw-scale effect sizes over 100 personalities. CS is measured in scale points. CR, PS, and MCQ are measured in percentage points. G, C, and M denote GPT-4o, Claude Sonnet 4.6, and Gemini 3.1 Pro. Positive values favor $A$ for CS, PS, and MCQ. Negative values favor $A$ for CR. $^{*}p<0.05$, $^{**}p<0.01$, and $^{***}p<0.001$.}
\label{tab:paired_significance}
\end{table*}

\section{Memorization and Novel-Topic Robustness}
\label{app:robustness}
We compare Memory-100 with Simple Prompt using two memorization diagnostics. First, we divide personalities into tertiles using training Q\&A count as a proxy for public exposure. Second, we divide test items around GPT-4.1's June 1, 2024 cutoff. The temporal split prevents the post-cutoff interviews themselves from appearing in pretraining, but those interviews may revisit facts discussed earlier. These diagnostics therefore test a memorization-only explanation without ruling out all contamination.

We also measure each test question's maximum cosine similarity to the same personality's training questions using text-embedding-3-small. We divide the 30,355 test questions into quartiles at 0.505, 0.590, and 0.679 similarity. Q1 contains the most novel questions and Q4 contains the most similar. Table~\ref{tab:robustness_results} reports Memory-100 minus Simple Prompt. Gains show no monotonic decline across popularity tiers. They also persist on post-cutoff items and in Q1 for CS and MCQ. CR is less consistent on novel questions, with small regressions under GPT-4o in Q1 and Q2 but improvements under Claude and Gemini.
\begin{table*}[t]
\centering
\small
\setlength{\tabcolsep}{3pt}
\begin{tabular}{@{}lccccccc@{}}
\toprule
\textbf{Subset}
& \multicolumn{3}{c}{\textbf{$\Delta$CS}}
& \multicolumn{3}{c}{\textbf{$\Delta$CR}}
& \textbf{$\Delta$MCQ} \\
\cmidrule(lr){2-4}\cmidrule(lr){5-7}
& G & C & M & G & C & M & Acc. \\
\midrule
\multicolumn{8}{l}{\textit{(a) Memorization diagnostics}} \\
Low popularity ($n=33$) & +0.26 & +0.37 & +0.37 & -1.01 & -4.94 & -2.91 & +1.71 \\
Mid popularity ($n=33$) & +0.21 & +0.36 & +0.29 & +0.39 & -5.14 & -3.24 & +1.41 \\
High popularity ($n=34$) & +0.24 & +0.40 & +0.31 & -0.84 & -4.46 & -3.52 & +1.91 \\
Pre-cutoff ($n=2{,}794$) & +0.52 & +0.61 & +0.70 & -5.62 & -8.48 & -7.98 & +1.69 \\
Post-cutoff ($n=27{,}561$) & +0.21 & +0.36 & +0.28 & +0.05 & -4.43 & -2.80 & +1.70 \\
\midrule
\multicolumn{8}{l}{\textit{(b) Train-test topic similarity}} \\
Q1, most novel & +0.15 & +0.24 & +0.14 & +0.75 & -1.34 & -1.81 & +1.39 \\
Q2 & +0.14 & +0.28 & +0.20 & +0.32 & -3.66 & -2.10 & +0.98 \\
Q3 & +0.21 & +0.34 & +0.29 & -0.28 & -4.40 & -2.60 & +0.98 \\
Q4, most similar & +0.45 & +0.67 & +0.65 & -2.69 & -9.79 & -6.61 & +3.44 \\
\bottomrule
\end{tabular}
\caption{Memory-100 minus Simple Prompt by diagnostic subset. G, C, and M denote GPT-4o, Claude Sonnet 4.6, and Gemini 3.1 Pro. Positive values favor Memory-100 for CS and MCQ. Negative values favor Memory-100 for CR. Popularity uses training Q\&A count as a proxy.}
\label{tab:robustness_results}
\end{table*}

\section{Automated Transcript Filtering}
\label{app:filtering_example}
\begin{figure*}[h]
\centering
\small
\begin{tcolorbox}[
    colback=white, 
    colframe=black, 
    title=\textbf{Task: Automated Interview Transcript Verification}, 
    width=\textwidth,
    sharp corners=south,
    boxrule=0.8pt
]

\begin{minipage}[t]{0.48\textwidth}
    \textbf{System Prompt} \\
    \textit{Instructs the model to classify transcripts:}
    
    You are an expert content analyst. Analyze interview transcripts to identify genuine interviews vs other content types. Pay special attention to: (1) number of distinct speakers, (2) presence of back-and-forth dialogue vs monologue, (3) title and description keywords. Reject entries where only one person speaks throughout. Respond with valid JSON only.

    \textbf{Acceptance Criteria:}
    \begin{itemize}
        \item Target personality is answering questions from an interviewer or host
        \item Back-and-forth conversational dialogue is present
        \item Target personality is the primary focus of the interview
    \end{itemize}

    \textbf{Rejection Criteria:}
    \begin{itemize}
        \item Target personality is the host or interviewer
        \item Group or panel discussions with shared focus
        \item Scripted monologues, tutorials, or reaction content
        \item Only one speaker detected throughout the transcript
    \end{itemize}
\end{minipage}
\hfill
\begin{minipage}[t]{0.48\textwidth}
    \textbf{User Prompt (Synthetic Example)} \\
    \textit{Placeholders shown in brackets are substituted at runtime:}
    
    Analyze this content to determine if \textbf{[PERSON\_NAME]} is being INTERVIEWED.

    TITLE: ``[PERSON\_NAME] on Life, Career, and New Album''
    
    TRANSCRIPT SAMPLE: ``\textit{Host: Welcome to the show. Let's start with your childhood. What was it like growing up? [PERSON\_NAME]: Oh, it was wonderful. I grew up in a small town and music was everything to my family. Host: When did you first know you wanted to be a singer? [PERSON\_NAME]: Probably around age twelve. I entered a local talent show and never looked back...}''
    
    \vspace{0.5em}
    \textbf{Expected Output:}

    \texttt{\{"is\_interview": true, "confidence": 0.95, "interview\_type": "target\_as\_interviewee", "speakers\_detected": 2, "has\_dialogue": true, "reason": "Two speakers with clear back-and-forth. Host asks questions, [PERSON\_NAME] provides personal answers."\}}
\end{minipage}

\vspace{0.5em}
\hrule
\vspace{0.5em}

\textbf{Filtering Logic.} Each transcript is assessed independently. Transcripts receiving \texttt{is\_interview: false} or a confidence score below the acceptance threshold are excluded. This automated stage removes group discussions, monologue-format content (e.g., scripted Q\&A where questions appear on screen), and entries where the target personality is the host rather than the interviewee.

\end{tcolorbox}
\caption{Worked example of the automated interview verification prompt used in the quality control pipeline. The LLM evaluates each transcript against structured acceptance and rejection criteria, outputting a JSON classification with confidence score. \texttt{[PERSON\_NAME]} and transcript content are substituted with actual values at runtime.}
\label{fig:filtering_example}
\end{figure*}

To ensure that only genuine one-on-one interviews are retained, we employ an LLM-based verification stage before human annotation. For each candidate transcript, GPT-4.1 receives the entry title, description, and a transcript sample, then classifies whether the target personality is being interviewed based on structured acceptance and rejection criteria. Figure~\ref{fig:filtering_example} presents a synthetic worked example illustrating the prompt format and expected output.

\section{Human Annotation}
\label{app:annotation}

\subsection{Annotation Guidelines}

To ensure the high quality of the \framework{} dataset, we provided human annotators with the following specific task card. The instructions in Figure~\ref{fig:annotation_card} explicitly define the acceptance criteria used to filter the dataset.
\begin{figure*}[h] 
\centering
\small  
\begin{tcolorbox}[
    colback=white, 
    colframe=black, 
    title=\textbf{Task: Interview Quality Annotation}, 
    width=\textwidth, 
    sharp corners=south,
    boxrule=0.8pt
]

\textbf{OBJECTIVE} \\
We need your help to verify the quality of interviews in our test set. Each row represents ONE interview. Your task is to determine if each video is a suitable source of dialogue data for the target personality.

\vspace{0.5em}
\hrule
\vspace{0.5em}

\begin{minipage}[t]{0.48\textwidth}
    \textbf{YES (Accept)} \\
    \textit{Mark YES only if the video meets \underline{ALL} conditions:}
    \begin{itemize}
        \item \textbf{Format:} Genuine interview format.
        \item \textbf{Identity:} Features the specified target personality.
        \item \textbf{Language:} Conversation is conducted in English.
        \item \textbf{Role:} Subject is the \textbf{interviewee} and is interviewed \textbf{alone} (1-on-1).
        \item \textbf{Substance:} Content is substantive (no brief greetings or promotional clips).
    \end{itemize}
\end{minipage}
\hfill
\begin{minipage}[t]{0.48\textwidth}
    \textbf{NO (Reject)} \\
    \textit{Mark NO if the video exhibits \underline{ANY} of these issues:}
    \begin{itemize}
        \item \textbf{Wrong Identity:} Person is not the target personality.
        \item \textbf{Language Mismatch:} Not in English.
        \item \textbf{Group Format:} Subject is interviewed with others (panels).
        \item \textbf{Invalid Format:} Music video, movie trailer, ad, or fan edit.
        \item \textbf{Role Mismatch:} Target personality is the host.
        \item \textbf{Poor Quality:} Audio is unintelligible.
    \end{itemize}
\end{minipage}

\vspace{1em}
\textbf{MAYBE (Unsure)} \\
Use sparingly. Only if the format is ambiguous (e.g., ``questions on screen'' format where subject answers prompts in a monologue).

\vspace{0.5em}
\hrule
\vspace{0.5em}

\textbf{REVIEW STRATEGY} \\
You do not need to watch the full video. Please follow these steps:
\begin{enumerate}
    \item Verify video title and thumbnail for obvious mismatches.
    \item Scrub through 4--5 different timestamps (beginning, middle, end).
    \item Listen to 3--5 seconds of audio at each point to confirm language and quality.
\end{enumerate}

\end{tcolorbox}
\caption{The exact annotation instructions provided to human workers for filtering the interview videos.}
\label{fig:annotation_card}
\end{figure*}

\subsection{Quality Assurance Process}
We followed a structured, multi-round quality assurance (QA) protocol to ensure annotation reliability, consisting of four stages:

\begin{enumerate}
    \item \textbf{Annotation (Round 1).} Annotators independently label each transcript according to the guidelines.
    \item \textbf{QA Review (Round 2).} A dedicated QA reviewer examines annotations, marking each as correct or incorrect with error type comments.
    \item \textbf{QA Evaluation (Round 3).} A secondary audit conducts sample-based verification of QA outputs to confirm review accuracy.
    \item \textbf{Acceptance (Round 4).} Final delivery validation by the project lead to ensure the data meets quality standards.
\end{enumerate}

\subsection{Risk-Based QA Coverage}
Given the volume of approximately 32,000 records, we employed a risk-based QA sampling strategy to maximize error detection while maintaining efficiency. QA coverage was adjusted dynamically based on each annotator's observed accuracy:

\begin{itemize}
    \item \textbf{Full QA coverage} for annotators with accuracy below 92\%, to prevent error leakage from higher-risk annotators.
    \item \textbf{Heavy sampling (30--50\%)} for annotators in the low-to-mid 90\% accuracy range, to validate stability and catch drift.
    \item \textbf{Light sampling (5--10\%)} for annotators above 95\% accuracy, with periodic spot checks.
\end{itemize}

Table~\ref{tab:annotator_accuracy} reports the anonymized per-annotator accuracy and the corresponding QA strategy applied.

\begin{table}[h]
\centering
\small
\begin{tabular}{lcc}
\toprule
\textbf{Annotator} & \textbf{Accuracy (\%)} & \textbf{QA Strategy} \\
\midrule
Annotator A & 97.78 & Light sampling \\
Annotator B & 95.99 & Light sampling \\
Annotator C & 93.99 & Heavy sampling \\
Annotator D & 93.75 & Heavy sampling \\
Annotator E & 93.57 & Heavy sampling \\
Annotator F & 86.39 & Full QA \\
Annotator G & 85.09 & Full QA \\
Annotator H & 83.12 & Full QA \\
\bottomrule
\end{tabular}
\caption{Per-annotator accuracy and QA coverage strategy. Accuracy is computed as the proportion of annotations confirmed correct during QA review.}
\label{tab:annotator_accuracy}
\end{table}

\section{Human Evaluation of Metrics}
\label{app:metrics_human_eval}

To validate the LLM-based evaluation metrics used in \framework{}, we conduct a human annotation study covering Content Similarity and Factual Consistency. Two trained annotators independently evaluate 300 items per metric. A pilot study of 20 items per task is conducted first, followed by disagreement discussion to calibrate understanding. All disagreements in the full study are resolved by a third-party adjudicator to produce a gold standard.

\paragraph{Content Similarity.} Annotators are presented with a question, the ground truth answer, and two generated responses (A and B) with different LLM-assigned scores. They select which response better matches the ground truth (A, B, or Tie), along with a confidence level (High, Medium, Low). Response pairs are sampled across score gaps of 1--4 to test the metric's discriminative power at varying difficulty levels. Annotator identities and LLM scores are hidden during annotation.

\paragraph{Factual Consistency.} Annotators are shown a generated response, the relevant facts about the personality, and the LLM judge's predicted label (Entailment, Neutral, or Contradiction) with its explanation. They judge whether the predicted label is correct (Yes, No, or Unsure), and if incorrect, provide the correct label. This design directly measures the LLM judge's classification accuracy.

\paragraph{Adjudication.} For Content Similarity, 91 disagreement cases (30.3\%) are adjudicated; for Factual Consistency, 50 cases (16.7\%). The adjudicator reviews both annotators' responses and provides a final label with written justification.

\paragraph{Upstream artifact audits.} Factual Consistency also depends on fact summaries extracted by GPT-4.1 from held-out interview Q\&As. A researcher checked 300 sampled summary statements against the source and found that 10 (3.3\%) contradicted it. MCQ scoring is exact-match, but item construction uses an LLM. To check item quality, a researcher viewed the source Q\&A and answered 300 sampled four-option MCQs. The selected answer matched the labeled gold option in 294 cases (98.0\%). The other six items were either answered differently or judged insufficiently supported by the source. These audits assess sampled construction quality.

\section{Thematic Question Categories}
\label{app:topic_categories}
\begin{table*}[h]
\centering
\small
\begin{tabular}{p{2.5cm} p{5cm} p{5cm}}
\toprule
\textbf{Category} & \textbf{Definition \& Sub-dimensions} & \textbf{Example Questions} \\
\midrule
\textit{Social Identity} &
Demographic and identity indicators including age, gender identity, race/ethnicity, nationality, education, occupation, marital status, political orientation, religion, and socioeconomic background. &
``Where are you from?'' \newline
``What is your educational background?'' \newline
``What is your political stance?'' \\
\midrule
\textit{Motivations and Values} &
Core beliefs and goals reflecting fundamental human values: benevolence, power, universalism, achievement, tradition, hedonism, stimulation, security, self-direction, and conformity. &
``What motivates you?'' \newline
``What is most important to you in life?'' \newline
``How do you define success?'' \\
\midrule
\textit{Identity Narrative} &
Personal and professional life stories including childhood memories, formative experiences, career journeys, key relationships, and how one's sense of self has evolved over time. &
``Tell me your life story.'' \newline
``What shaped who you are today?'' \newline
``Walk me through your career.'' \\
\midrule
\textit{Psychological Traits} &
Behavioral tendencies and personality dimensions aligned with the Big Five model: Extraversion, Agreeableness, Conscientiousness, Neuroticism, and Openness to Experience. &
``Are you more of an introvert or extrovert?'' \newline
``How do you handle stress?'' \newline
``Do you trust people easily?'' \\
\bottomrule
\end{tabular}
\caption{Definitions, sub-dimensions, and example questions for the four thematic categories used in Q\&A pairs.}
\label{tab:topic_categories}
\end{table*}
\begin{table}[t]
\centering
\small
\begin{tabular}{p{0.95\columnwidth}}
\toprule
\textbf{Raw Transcript Segment} \\
\midrule
\textit{``...so um tell me about your you know your early days what was it like growing up in that environment and how did that shape you as a person I mean you've talked about this before but... well it was uh it was tough honestly we didn't have much my mom worked two jobs and I think that's where I got my work ethic from you know seeing her get up at 5am every single day that never left me...''} \\
\midrule
\textbf{Structured Output} \\
\midrule
\textbf{Speaker Attribution:} \\
\textsc{Host:} So tell me about your early days, what was it like growing up in that environment and how did that shape you as a person? \\[0.3em]
\textsc{[Person\_Name]:} Well, it was tough honestly. We didn't have much. My mom worked two jobs and I think that's where I got my work ethic from. Seeing her get up at 5am every single day, that never left me. \\[0.5em]
\textbf{Extracted Q\&A Pair:} \\
\textbf{Q:} What was it like growing up and how did that environment shape you? \\
\textbf{A:} It was tough. We didn't have much. My mom worked two jobs and I think that's where I got my work ethic from. Seeing her get up at 5am every single day, that never left me. \\[0.3em]
\textbf{Topic:} Identity Narrative \\
\bottomrule
\end{tabular}
\caption{Example of dialogue structuring output. A raw transcript segment is processed into a structured Q\&A pair with speaker attribution, disfluency removal, and topic classification.}
\label{tab:qa_example}
\end{table}
Each question-answer pair is classified into one of four thematic categories derived from established frameworks in personality and identity research~\cite{mcadams2013redemptive,schwartz2012overview, venkit2026need}. Table~\ref{tab:topic_categories} provides detailed definitions, sub-dimensions, and example questions for each category. Table~\ref{tab:qa_example} illustrates the full dialogue structuring pipeline on a synthetic transcript segment, showing speaker attribution, disfluency removal, and topic classification.

\section{Dataset Distribution Analysis}
\label{app:distribution}

This appendix provides an analysis of the distribution of interview data per subject, focusing on total video duration and the volume of extracted Q\&A pairs.

\subsection{Interview Video Duration Distribution}
\label{app:duration_dist}
\begin{figure*}[h]
    \centering
    \includegraphics[width=\textwidth]{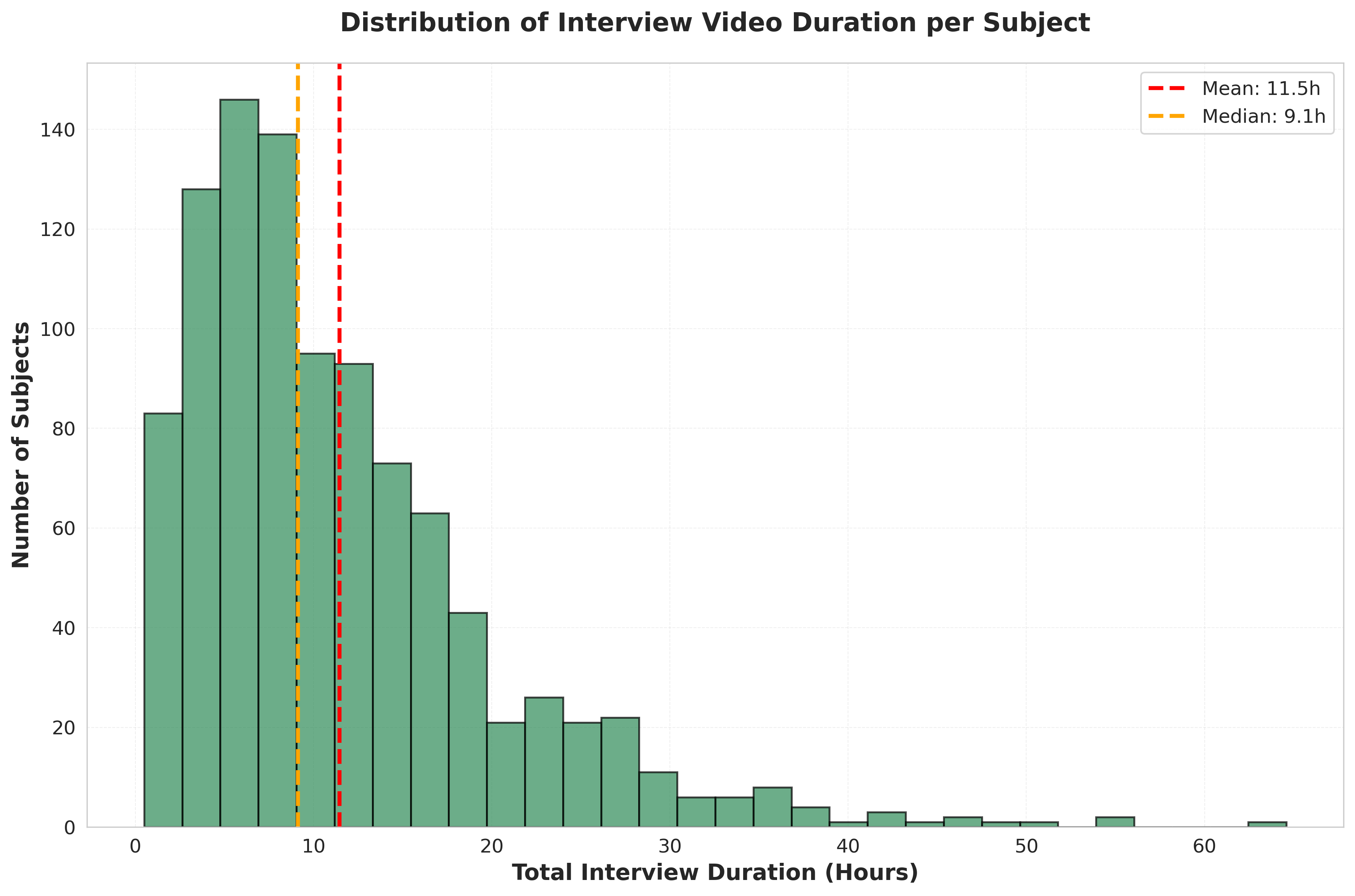} 
    \caption{Distribution of total interview video duration per subject.}
    \label{fig:duration_dist}
\end{figure*}

Figure~\ref{fig:duration_dist} presents the distribution of total interview video duration across the 1,000 subjects in our dataset. The distribution indicates balanced coverage, with the majority of subjects possessing between 5--15 hours of content. The dataset demonstrates a mean duration of 11.46 hours and a median of 9.11 hours per subject. The relatively symmetric nature of the distribution reflects natural variations in the availability of interview content for different public figures.

The dataset spans a wide range of durations, from a minimum of 0.49 hours to a maximum of 64.59 hours per subject. The total aggregated duration across all 1,000 subjects reaches 11,464.30 hours, equivalent to approximately 478 days of continuous video content. As illustrated in the figure, the dataset successfully captures subjects with varying levels of media presence, from those with limited but sufficient content to highly prominent figures with extensive histories. This variation enhances the dataset's representativeness and generalizability across different levels of public visibility.

\subsection{Q\&A Pair Count Distribution}
\label{app:qa_dist}
\begin{figure}[t]
    \centering
    \includegraphics[trim={0 0 0 0},clip,width=\textwidth]{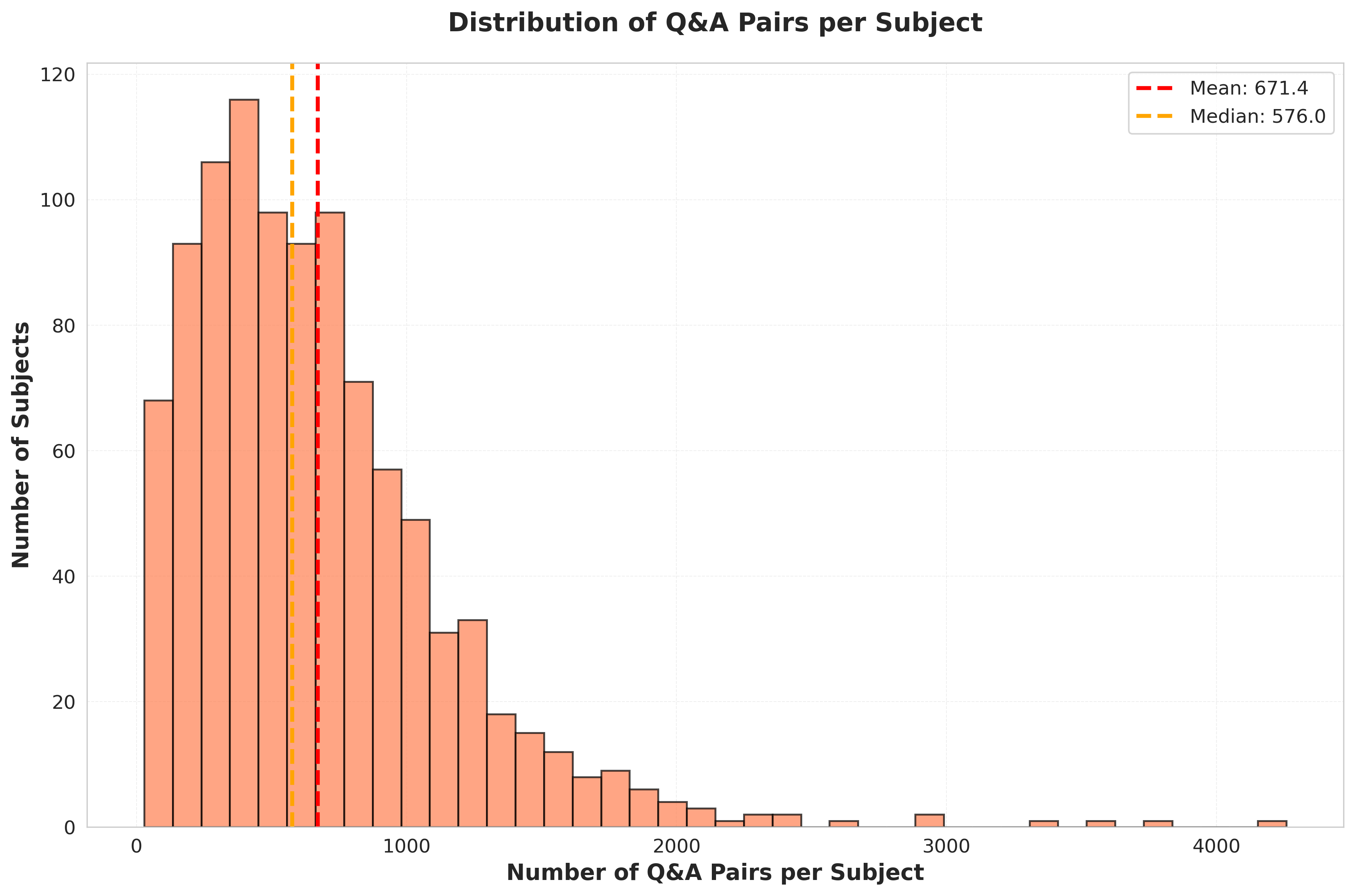}
    \caption{Distribution of Q\&A pair counts per subject.}
    \label{fig:qa_count_dist}
\end{figure}

Figure~\ref{fig:qa_count_dist} displays the distribution of Q\&A pair counts per subject. The distribution exhibits a right-skewed pattern, with a mean of 671.4 pairs and a median of 576 pairs per subject. The total dataset comprises 671,424 Q\&A pairs, providing a comprehensive foundation for personality assessment.

The volume of pairs ranges from a minimum of 28 to a maximum of 4,259 per subject. The upper tail demonstrates that some subjects contribute substantially more data, reflecting both the volume of their media appearances and the conversational density of their interviews. Most subjects fall within the 300--900 Q\&A pair range. The right-skewed nature of the distribution is expected, as public personalities vary significantly in interview frequency and career longevity. This natural variation allows for the evaluation of personality modeling approaches under diverse data availability conditions, ranging from data-scarce to data-rich scenarios.

\section{Generation Methods: Technical Details}
\label{app:methods}

This appendix provides detailed mathematical formulations and analysis for the generation methods described in Section~\ref{sec:experiments}.

\subsection{Context Size Analysis}

For each method, we provide token count estimates and computational cost analysis.

\paragraph{Simple Prompt:}
\[
\text{tokens}(\mathcal{P}_{\text{simple}}) \approx 30 + |q| \approx 50 \text{ tokens}
\]
where $|q|$ is the question length (average 15-20 tokens).

\paragraph{Wiki-based:}
\[
\scriptsize
\text{tokens}(\mathcal{P}_{\text{wiki}}) \approx 30 + |\text{profile}(c)| + |q| \approx 200\text{-}500 \text{ tokens}
\]
where $|\text{profile}(c)|$ ranges from 150-450 tokens depending on Wikipedia entry length.

\paragraph{Wiki-Long:}
\[
\text{tokens}(\mathcal{P}_{\text{wiki-long}}) \approx 30 + |\text{excerpt}(c)| + |q| \approx 2{,}500 \text{ tokens}
\]
The stored Wikipedia excerpt is capped at 10,000 characters (9,856 characters on average). Of the 100 evaluated personalities, 97 reach this cap. Wiki-Long therefore provides roughly seven times the Wiki-based context, but remains shorter than Chronological-100 and retains Wikipedia's third-person format.

\paragraph{Chronological-based:}
\[
\text{tokens}(\mathcal{P}_{\text{chrono}}) \approx 30 + \sum_{i=1}^{m} (|q_i| + |a_i|) + |q|
\]
With average Q\&A pair lengths of 80-100 tokens: $m = 100$ yields $\sim$10k tokens, $m = 500$ yields $\sim$45k tokens, $m = 1000$ yields $\sim$90k tokens.

\paragraph{Memory-based:}
\[
\scriptsize
\text{tokens}(\mathcal{P}_{\text{memory}}) \approx 30 + \sum_{i=1}^{k} (|q_i| + |a_i|) + |q| \approx 10\text{k tkn}
\]
With $k = 100$ retrieved examples (similar to chronological-based with 100 examples).

\subsection{Memory-Based Retrieval Mechanism}

We provide the complete mathematical formulation for the retrieval process used in the memory-based method.

\subsubsection{Embedding Function}

Given a text $t$ (question or answer), we compute a dense vector representation $\mathbf{e}_t \in R^d$ using a pre-trained embedding model:

\[
\mathbf{e}_t = \phi(t)
\]

where $\phi$ is the text-embedding-3-small model with embedding dimension $d = 1536$. The embeddings are L2-normalized such that $\|\mathbf{e}_t\| = 1$.

\subsubsection{Similarity Computation}

The semantic similarity between two texts $t_1$ and $t_2$ is computed via cosine similarity of their embeddings:

\[
\text{sim}(t_1, t_2) = \frac{\phi(t_1) \cdot \phi(t_2)}{\|\phi(t_1)\| \|\phi(t_2)\|} = \mathbf{e}_{t_1} \cdot \mathbf{e}_{t_2}
\]

For normalized embeddings, this simplifies to the dot product. The similarity score ranges from $-1$ to $1$, with higher values indicating greater semantic similarity.

\subsubsection{Top-k Retrieval}

Given a test question $q$ and a personality's training dataset $\mathcal{D}_c = \{(q_i, a_i)\}_{i=1}^{N_c}$ of $N_c$ Q\&A pairs, we retrieve the $k$ most similar training questions:

\[
\mathcal{R}_k(q, \mathcal{D}_c) = \underset{\mathcal{S} \subseteq \mathcal{D}_c, |\mathcal{S}| = k}{\arg\max} \sum_{(q_i, a_i) \in \mathcal{S}} \text{sim}(q, q_i)
\]

Operationally, we:
\begin{enumerate}
    \item Compute similarity scores: $s_i = \text{sim}(q, q_i)$ for all $i \in \{1, \ldots, N_c\}$
    \item Sort training examples by descending similarity: $s_{\sigma(1)} \geq s_{\sigma(2)} \geq \cdots \geq s_{\sigma(N_c)}$
    \item Select the top-$k$: $\mathcal{R}_k(q, \mathcal{D}_c) = \{(q_{\sigma(i)}, a_{\sigma(i)})\}_{i=1}^{k}$
\end{enumerate}

\subsubsection{Computational Complexity}

The retrieval process has two phases:

\paragraph{Pre-computation (once per personality):}
\begin{itemize}
    \item Embed all training questions: $O(N_c)$ API calls to embedding model
    \item Store embeddings: $O(N_c \cdot d)$ space
\end{itemize}

\paragraph{Query-time (per test question):}
\begin{itemize}
    \item Embed test question: $O(1)$ API call
    \item Compute similarities: $O(N_c \cdot d)$ floating-point operations (dot products)
    \item Sort and select top-$k$: $O(N_c \log N_c)$ comparisons
    \item Total: $O(N_c \cdot d + N_c \log N_c)$ per test question
\end{itemize}

For our dataset with $N_c \approx 1000$ training examples per personality, $d = 1536$, and $k = 100$ retrieved examples, query-time retrieval takes $\sim$0.1-0.2 seconds per test question on standard hardware.

\subsubsection{Random Baseline}

The random selection baseline samples $k$ examples uniformly at random:

\[
\mathcal{R}_{\text{random}}(q, \mathcal{D}_c, k) = \text{UniformSample}(\mathcal{D}_c, k)
\]

To ensure reproducibility, we use deterministic randomness based on the test question's identifier: $\text{seed} = \text{hash}(\text{qa\_id}) \mod 2^{32}$.

This baseline isolates the effect of relevance-based selection versus simply having diverse examples in context.

\subsubsection{Expected Retrieval Quality}

Relevance-based retrieval should yield higher average similarity than random selection. Formally, let $\bar{s}_k$ denote the average similarity of the $k$ retrieved examples:

\[
\bar{s}_k = \frac{1}{k} \sum_{i=1}^{k} s_{\sigma(i)}
\]

We expect:
\[
\mathbb{E}[\bar{s}_k^{\text{relevance}}] > \mathbb{E}[\bar{s}_k^{\text{random}}]
\]

This hypothesis is validated by comparing the performance of relevance-based versus random selection in our experiments.

\section{Full Dataset Validation}
\label{app:full_validation}

To confirm that our main findings generalize beyond the 100 data-rich personalities, we evaluate three methods on all 1,000 personalities. The 100-personality subset used in the main experiments averages 1,375 training Q\&A pairs per personality (minimum 1,001), while the remaining 900 average 437 (minimum 10, median 412). Memory-based retrieval is excluded from this evaluation as pre-computed embeddings were generated only for the 100-personality subset.

\begin{table}[t]
\centering
\small
\begin{tabular}{lccccc}
\toprule
\textbf{Method} & \textbf{CS} & \textbf{CR} & \textbf{PS} & \textbf{MCQ-A} & \textbf{MCQ-R} \\
 & (1-5) & (\%) & (\%) & (\%) & \\
\midrule
Simple Prompt & 3.27 & 9.00 & 72.6 & 86.3 & 0.764 \\
Wiki-based & 3.27 & \textbf{8.48} & 70.2 & 86.3 & 0.764 \\
Chrono-based & \textbf{3.43} & 8.53 & \textbf{76.5} & \textbf{87.9} & \textbf{0.789} \\
\bottomrule
\end{tabular}
\caption{Performance of three methods on the full 1000-personality dataset. Chronological-based uses 100 examples.}
\label{tab:full_dataset_validation}
\end{table}

Table~\ref{tab:full_dataset_validation} shows that Chrono-100 achieves the highest Content Similarity, Personality Similarity, and MCQ performance on the full dataset, while Wiki-based has the lowest Contradiction Ratio by 0.05 points. Relative to the same three methods on the 100-personality subset, the Personality Similarity and MCQ orderings are preserved, but the Content Similarity and Contradiction Ratio orderings change. Contradiction ratios increase from 6.10\%--7.10\% on the 100-personality subset to 8.48\%--9.00\% on the full dataset.

\section{Contradiction Examples by Question Category}
\label{app:contradiction_examples}

Table~\ref{tab:contradiction_examples} presents representative contradiction errors from the memory-based method, one per question category. All examples are anonymized with identifiers replaced by placeholders. These examples illustrate how contradiction patterns differ across categories: Social Identity errors involve precise numerical facts, Identity Narrative errors fabricate biographical events, Motivations and Values errors invert stated beliefs, and Psychological Traits errors misrepresent behavioral tendencies.

\begin{table}[t]
\centering
\small
\setlength{\tabcolsep}{3pt}
\begin{tabular}{@{}p{1.6cm} p{2.4cm} p{3.0cm} p{3.0cm} p{1.8cm}@{}}
\toprule
\textbf{Category} & \textbf{Question} & \textbf{Ground Truth} & \textbf{Generated Response} & \textbf{Error Type} \\
\midrule
Social Identity &
How much [substance] do you consume in a day? &
``I consume about 800 milligrams a day.'' &
``I typically consume somewhere between 100 and 250 milligrams per day.'' &
Numerical fact (4$\times$ underestimate) \\
\midrule
Identity Narrative &
What are you doing to celebrate your [milestone] birthday? &
``I'll gather hundreds of my friends and maybe they'll roast me.'' &
``I'll probably have dinner with my family and a few close friends.'' &
Event fabrication (hundreds $\rightarrow$ few) \\
\midrule
Motivations \& Values &
Do you feel like you have to explain yourself to [group]? &
``I still feel like I have to explain myself.'' &
``No, I don't feel like I have to explain myself \ldots if you don't get it, you're probably not my people.'' &
Belief inversion \\
\midrule
Psych.\ Traits &
How do you feel about receiving praise from fans? &
``Praise is not---given the way my personality is built, this line is not working for me.'' &
``Oh, it's lovely. It's just really nice. It feels like being a dog and someone's giving you a treat.'' &
Trait reversal \\
\bottomrule
\end{tabular}
\caption{Representative contradiction examples by question category (memory-based method, anonymized). Each row shows the question, ground truth excerpt, generated response excerpt, and the nature of the contradiction.}
\label{tab:contradiction_examples}
\end{table}

\section{MCQ Robustness Analyses}
\label{app:mcq_robustness}

This appendix provides detailed analysis for the MCQ evaluation. We analyze robustness across training-data availability, distractor type, shuffled answer position, and answer-length results.

\paragraph{Performance Across Training-Data Regimes.}
Tables~\ref{tab:mcq_3opt_by_data} and~\ref{tab:mcq_4opt_by_data} report MCQ accuracy and reward stratified into ten groups based on the number of available training Q\&A pairs. Chronological-based prompting consistently achieves higher or comparable accuracy and uniformly higher reward across all data regimes in both 3-option and 4-option settings. Gains are present even in low-data groups, indicating that interview grounding improves factual calibration rather than merely increasing coverage. Improvements remain stable as training data scales, suggesting robustness across sparse and high-resource personalities.

\paragraph{Distractor-Type Selection.}
Tables~\ref{tab:mcq_3opt_distribution} and~\ref{tab:mcq_4opt_distribution} break down predictions by option type. Chronological-based prompting substantially reduces selection of the opposite/negation distractor in both 3-option and 4-option formats, while maintaining or improving overall accuracy. Because opposite selections incur the largest penalty in the reward metric, this reduction explains the clearer separation in reward than in raw accuracy.

\paragraph{Position Bias.}
Tables~\ref{tab:mcq_position_3opt} and~\ref{tab:mcq_position_4opt} report accuracy by shuffled answer position. All methods exhibit mild position effects, with earlier options selected more frequently. However, chronological-based prompting shows narrower performance variance across positions, indicating improved calibration under randomized distractor ordering.

\paragraph{Length Bias.}
Table~\ref{tab:mcq_length_bias} evaluates accuracy conditioned on whether the correct option is the longest answer. All methods display moderate length bias, with higher accuracy when the correct option is longest. Chronological-based prompting slightly reduces this gap but does not eliminate it. 

Overall, these analyses demonstrate that the MCQ improvements from chronological interview grounding are consistent across data regimes and are not artifacts of distractor position or answer-length biases.

\begin{table*}[t]
\centering
\small
\begin{tabular}{ccccccc}
\toprule
\textbf{Group} & \textbf{Avg Train} & \multicolumn{2}{c}{\textbf{Simple}} & \multicolumn{2}{c}{\textbf{Chronological}} \\
 & \textbf{Examples} & Acc (\%) & Reward & Acc (\%) & Reward \\
\midrule
1 & 74 & 87.0 & 0.774 & 88.2 & 0.796 \\
2 & 174 & 86.0 & 0.760 & 87.3 & 0.780 \\
3 & 258 & 85.5 & 0.751 & 87.3 & 0.779 \\
4 & 339 & 85.6 & 0.752 & 87.0 & 0.776 \\
5 & 412 & 83.0 & 0.706 & 84.4 & 0.729 \\
6 & 493 & 85.7 & 0.755 & 87.0 & 0.775 \\
7 & 590 & 84.7 & 0.734 & 85.9 & 0.752 \\
8 & 718 & 86.0 & 0.758 & 87.7 & 0.786 \\
9 & 873 & 86.1 & 0.761 & 87.5 & 0.783 \\
10 & 1375 & 85.6 & 0.749 & 87.1 & 0.773 \\
\midrule
\textbf{All} & \textbf{531} & \textbf{85.4} & \textbf{0.748} & \textbf{86.9} & \textbf{0.771} \\
\bottomrule
\end{tabular}
\caption{3-option MCQ accuracy and reward by training-data group.}
\label{tab:mcq_3opt_by_data}
\end{table*}

\begin{table*}[t]
\centering
\small
\begin{tabular}{ccccccc}
\toprule
\textbf{Group} & \textbf{Avg Train} & \multicolumn{2}{c}{\textbf{Simple}} & \multicolumn{2}{c}{\textbf{Chronological}} \\
 & \textbf{Examples} & Acc (\%) & Reward & Acc (\%) & Reward \\
\midrule
1 & 74 & 86.4 & 0.768 & 87.8 & 0.792 \\
2 & 174 & 85.7 & 0.757 & 87.4 & 0.785 \\
3 & 258 & 85.6 & 0.754 & 86.9 & 0.775 \\
4 & 339 & 85.4 & 0.753 & 86.4 & 0.770 \\
5 & 412 & 83.0 & 0.710 & 84.2 & 0.729 \\
6 & 493 & 86.1 & 0.763 & 86.8 & 0.774 \\
7 & 590 & 84.6 & 0.736 & 85.9 & 0.756 \\
8 & 718 & 86.0 & 0.763 & 87.4 & 0.785 \\
9 & 873 & 86.1 & 0.763 & 87.1 & 0.779 \\
10 & 1375 & 85.8 & 0.757 & 87.1 & 0.776 \\
\midrule
\textbf{All} & \textbf{531} & \textbf{85.5} & \textbf{0.752} & \textbf{86.7} & \textbf{0.771} \\
\bottomrule
\end{tabular}
\caption{4-option MCQ accuracy and reward by training-data group.}
\label{tab:mcq_4opt_by_data}
\end{table*}

\begin{table}[t]
\centering
\small
\begin{tabular}{lcccc}
\toprule
\textbf{Method} & Correct & Opposite & Near-Miss & Misconception \\
\midrule
Simple Prompt & 85.5\% & 6.0\% & 5.7\% & 2.8\% \\
Chrono-based & 86.7\% & 5.8\% & 5.0\% & 2.4\% \\
\bottomrule
\end{tabular}
\caption{4-option prediction distribution. Chronological-based reduces high-severity opposite selections.}
\label{tab:mcq_4opt_distribution}
\end{table}

\begin{table}[t]
\centering
\small
\begin{tabular}{lccc}
\toprule
\textbf{Method} & Pos A & Pos B & Pos C \\
\midrule
Simple Prompt & 89.7\% & 85.0\% & 81.7\% \\
Chrono-based & 89.8\% & 86.6\% & 84.3\% \\
\bottomrule
\end{tabular}
\caption{3-option per-position accuracy. Chronological-based reduces positional spread.}
\label{tab:mcq_position_3opt}
\end{table}

\begin{table}[t]
\centering
\small
\begin{tabular}{lcccc}
\toprule
\textbf{Method} & Pos A & Pos B & Pos C & Pos D \\
\midrule
Simple Prompt & 88.5\% & 85.7\% & 84.1\% & 82.4\% \\
Chrono-based & 88.6\% & 87.0\% & 85.9\% & 84.4\% \\
\bottomrule
\end{tabular}
\caption{4-option per-position accuracy. Chronological-based narrows positional variance.}
\label{tab:mcq_position_4opt}
\end{table}

\begin{table}[t]
\centering
\small
\begin{tabular}{lccc}
\toprule
\textbf{Method} & Longest Correct & Not Longest & Delta Acc \\
\midrule
Simple Prompt & 88.2\% & 80.2\% & +8.0\% \\
Wiki-based & 88.3\% & 80.2\% & +8.1\% \\
Chrono-based & 88.0\% & 81.0\% & +7.0\% \\
\bottomrule
\end{tabular}
\caption{3-option length bias. Chronological-based slightly reduces length advantage.}
\label{tab:mcq_length_bias}
\end{table}

\end{document}